%% file: main.tex
\documentclass{article} 
\usepackage{arxiv,times}

\input{math_commands.tex}

\usepackage{hyperref}
\usepackage{url}
\usepackage{graphicx}
\usepackage{float}
\usepackage{wrapfig}
\usepackage{booktabs}
\usepackage{multirow}
\usepackage{colortbl}
\usepackage[most]{tcolorbox}
\usepackage{enumitem}
\usepackage[T1]{fontenc}  
\usepackage[table]{xcolor}
\usepackage{tabularx}
\usepackage{longtable}

\definecolor{oursblue}{RGB}{239,246,252}
\definecolor{stdblue}{RGB}{74,119,175}
\definecolor{stdbluebg}{RGB}{228,239,250}
\definecolor{gainblue}{RGB}{106,157,198}

\newcommand{\ourscell}[1]{\cellcolor{oursblue}#1}

\definecolor{promptbluebg}{HTML}{EDF6FC}
\definecolor{promptblueframe}{HTML}{78ADD0}
\definecolor{promptbluetitle}{HTML}{E3F1FA}
\definecolor{promptpinkbg}{HTML}{FCEFF2}
\definecolor{promptpinkframe}{HTML}{D79AA6}
\definecolor{promptpinktitle}{HTML}{F8E4E9}

\newtcolorbox{vanillabox}[1]{
  enhanced,
  breakable,
  colframe=gray!55,
  colback=white,
  colbacktitle=gray!20,
  coltitle=black,
  fonttitle=\bfseries\small,
  fontupper=\footnotesize,
  boxrule=0.65pt,
  arc=2pt,
  left=5pt,
  right=5pt,
  top=4pt,
  bottom=4pt,
  before skip=6pt,
  after skip=6pt,
  title={#1},
  title after break={#1 (continued)}
}

\newtcolorbox{selfrevisionbox}[1]{
  enhanced,
  breakable,
  colframe=promptblueframe,
  colback=white,
  colbacktitle=promptbluetitle,
  coltitle=black,
  fonttitle=\bfseries\small,
  fontupper=\footnotesize,
  boxrule=0.65pt,
  arc=2pt,
  left=5pt,
  right=5pt,
  top=4pt,
  bottom=4pt,
  before skip=6pt,
  after skip=6pt,
  title={#1},
  title after break={#1 (continued)}
}

\newtcolorbox{lookplangraphbox}[1]{
  enhanced,
  breakable,
  colframe=orange!75!red,
  colback=white,
  colbacktitle=orange!15,
  coltitle=black,
  fonttitle=\bfseries\small,
  fontupper=\footnotesize,
  boxrule=0.65pt,
  arc=2pt,
  left=5pt,
  right=5pt,
  top=4pt,
  bottom=4pt,
  before skip=6pt,
  after skip=6pt,
  title={#1},
  title after break={#1 (continued)}
}

\newtcolorbox{causgroundbox}[1]{
  enhanced,
  breakable,
  colframe=violet!70,
  colback=white,
  colbacktitle=violet!15,
  coltitle=black,
  fonttitle=\bfseries\small,
  fontupper=\footnotesize,
  boxrule=0.65pt,
  arc=2pt,
  left=5pt,
  right=5pt,
  top=4pt,
  bottom=4pt,
  before skip=6pt,
  after skip=6pt,
  title={#1},
  title after break={#1 (continued)}
}

\usepackage{fancyhdr}

\fancypagestyle{preprintfirst}{%
  \fancyhf{}
  \fancyhead[L]{%
    \includegraphics[
      height=12pt,
      keepaspectratio
    ]{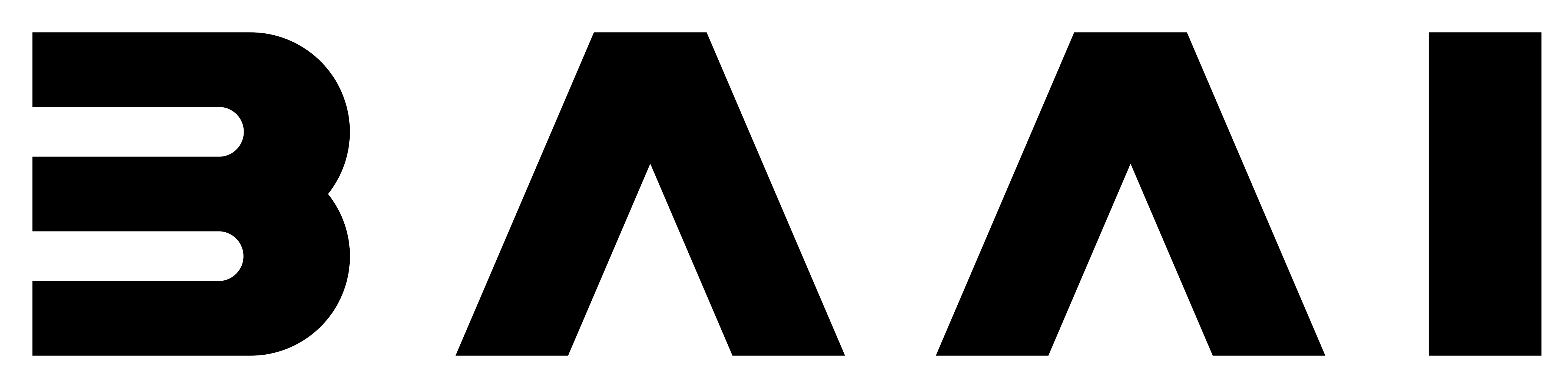}%
  }
  \fancyhead[R]{\small Preprint}
  \fancyfoot[C]{\thepage}
  \renewcommand{\headrulewidth}{0.4pt}
}

\fancypagestyle{preprintrest}{%
  \fancyhf{}
  \fancyhead[R]{\small Preprint}
  \fancyfoot[C]{\thepage}
  \renewcommand{\headrulewidth}{0.4pt}
}

\title{CausGround: Causal Interaction Grounding for Adaptive Embodied Planning}

\author{%
  \parbox[t]{\dimexpr\textwidth-2\tabcolsep\relax}{%
    \centering
    {\normalsize\bfseries
      Yijun Shen$^{1,2}$ \quad
      Diqi He$^{2,3}$ \quad
      Jinbang Guo$^{2}$ \quad
      Wanyue Zhang$^{2,4,5}$ \quad
      Jiaming Mi$^{1}$ \\
      Huan Yang$^{2}$ \quad
      Zhongyuan Wang$^{2}$ \quad
      Zhiwei Yu$^{2,\ast}$ \quad
      Kai Zhang$^{1,\ast}$
    }\\[6pt]
    {\small\normalfont
      $^{1}$East China Normal University \quad
      $^{2}$Beijing Academy of Artificial Intelligence (BAAI)\\[2pt]
      $^{3}$Northwestern Polytechnical University\\[2pt]
      $^{4}$School of Artificial Intelligence,
      University of Chinese Academy of Sciences\\[2pt]
      $^{5}$Institute of Automation,
      Chinese Academy of Sciences
    }%
  }%
}

\iclrfinalcopy 
\begin{document}

\maketitle

\pagestyle{preprintrest}
\thispagestyle{preprintfirst}

\begingroup
\renewcommand{\thefootnote}{\fnsymbol{footnote}}

\footnotetext[1]{Corresponding author.}

\endgroup

\input{Content/abstract}
\input{Content/intro}

\input{Content/related}
\input{Content/methods}
\input{Content/experiments}
\input{Content/conclusion}

\subsection*{AI use statement}
In this work, we used AI tools for manuscript language refinement and to assist the construction of the adapted VirtualHome visual benchmark.
Specifically, AI-based image editing was used to repair RGB observations that were visually inconsistent with the underlying simulator states, as described in Appendix~\ref{app:virtualhome_repair}.
The repaired observations and AI-assisted manuscript edits were reviewed by the authors, who take full responsibility for the final content of this work.

\subsection*{Reproducibility statement}
We provide detailed benchmark construction and evaluation protocols in Appendix~\ref{app:environment}, additional methodological details of CausGround in Appendix~\ref{app:method}, and matched experimental settings, baseline implementations, library construction, retrieval and grounding procedures, and prompt templates in Appendix~\ref{app:implementation}. 
Additional ablation studies and real-world deployment details are provided in Appendices~\ref{app:additional_ablations} and~\ref{app:real_world}, respectively.

\bibliography{iclr2027_conference}
\bibliographystyle{iclr2027_conference}

\input{Content/appendix}

\end{document}

%% file: math_commands.tex
\usepackage{amsmath,amsfonts,bm}

\def\eqref#1{equation~\ref{#1}}

\def\1{\bm{1}}

\DeclareMathAlphabet{\mathsfit}{\encodingdefault}{\sfdefault}{m}{sl}
\SetMathAlphabet{\mathsfit}{bold}{\encodingdefault}{\sfdefault}{bx}{n}



%% file: Content/abstract.tex
\begin{abstract}
Multimodal large language models (MLLMs) have demonstrated promising reasoning capabilities, but deploying them for physical interaction requires reliable planning that accounts for when actions are executable and how they change the world.
These preconditions and effects are often environment-dependent and are not fully captured by MLLMs, making adaptation to new environments challenging. 
Yet existing adaptation approaches either update model parameters through costly environment-specific training or rely on unstructured interaction experience that is difficult to ground and reuse.
To address this challenge, we introduce~\textbf{CausGround}, a non-parametric causal interaction grounding framework that distills interaction experience into reusable action-centered causal knowledge.
CausGround learns from experience by identifying action preconditions and effects, and organizes them in a factorized world-state representation that jointly captures object attributes and inter-object relations.
At planning time, task-relevant causal dependencies are grounded in the current observation, supporting backward reasoning over action preconditions and forward reasoning over action effects for action selection and multi-step causal planning.
We demonstrate the effectiveness of CausGround across diverse long-horizon embodied settings spanning simulated high-level planning, interaction reasoning, and real-world robot manipulation.
Evaluated on the adapted VirtualHome and RoboCasa benchmarks, CausGround consistently improves vision-based high-level planning across MLLM backbones, improving average task success from 24.9\% to 53.3\% on VirtualHome and closed-loop planning success from 43.0\% to 57.0\% on RoboCasa.
Consistent gains are also observed on SWITCH benchmark, with average interaction reasoning accuracy improving from 25.8\% to 31.9\%, indicating stronger reasoning over action selection, recovery, and verification.
Moreover, real-world experiments further show that CausGround can reason backward from MLLM-interpreted goal states to derive executable high-level action sequences for sequential manipulation.
These results demonstrate that explicit action preconditions and effects provide a reusable causal grounding mechanism for long-horizon embodied planning across diverse environments without parameter updates.

\textbf{Project Page}: \url{https://causground.github.io}
\end{abstract}

%% file: Content/intro.tex
\section{Introduction}
\label{sec:intro}
\begin{figure*}[t]
    \centering
    \includegraphics[width=0.90\textwidth]{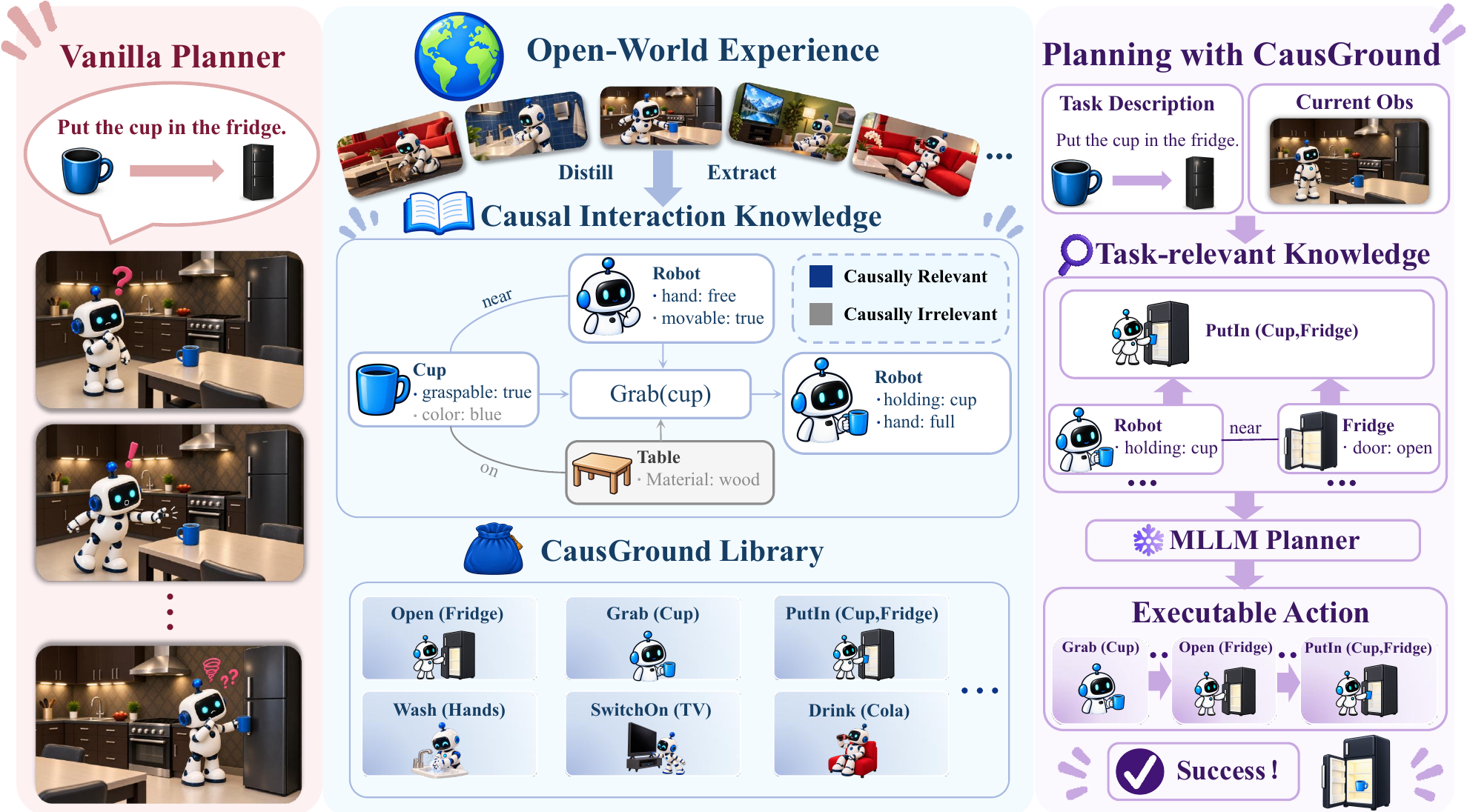}
    \caption{
        \textbf{Overview of CausGround.} CausGround distills open-world interaction experience into reusable action-centered causal knowledge by identifying action preconditions and effects. At planning time, task-relevant causal dependencies are grounded in the current observation to form a local causal graph, supporting backward reasoning over preconditions and forward reasoning over effects for executable long-horizon planning without parameter updates.
    }
    \label{fig:teaser}
\end{figure*}

Embodied agents powered by multimodal large language models (MLLMs) have demonstrated promising reasoning capabilities for high-level planning~\citep{chen2025planning, driess2023palm, feng2025reflective}. \textbf{Yet successful embodied interaction requires more than knowing what to do: agents must know when an action is executable and how it changes the world.} Such interaction knowledge is often environment-dependent, involving available actions, object states, action preconditions, and effects~\citep{kim2025multi, han2024interpret, liu2024learning}. As illustrated in Fig.~\ref{fig:teaser}, a planner may correctly infer that an object should be placed into a fridge, yet attempt the action before satisfying the precondition that the fridge be open. Such precondition violations can accumulate over long-horizon execution, leading to invalid actions, repeated attempts, and costly trial-and-error~\citep{raman2024cape, zhi2025closed, chang2025partnr}. Effective embodied planning therefore requires agents to account for task semantics while adapting their planning to the interaction dynamics of the environment in which the task is executed.

Existing approaches still fall short in enabling efficient environment adaptation. A direct strategy is to specialize embodied agents through fine-tuning or reinforcement learning~\citep{xu2026roboagent, zhai2024fine, feng2025towards}, but such parametric adaptation requires substantial environment-specific interaction data and repeated optimization, making it costly to deploy across diverse environments, particularly in real-world settings. 
Non-parametric approaches instead incorporate interaction feedback or episodic memory at inference time~\citep{anwar2025remembr}, but such experience is often retained as unstructured trajectories or task-level memories~\citep{ginting2025enter, buoso2025select2plan}. This makes it difficult to identify which parts of an experience are causally relevant rather than incidental, reuse them across tasks, or ground them in the current visual scene. 
Moreover, existing representations do not always jointly capture the fine-grained object attributes and inter-object relations required to determine whether an action's preconditions are satisfied~\citep{chen2025planning, shen2026reward}.
This raises a central question: Can interaction experience be distilled into reusable causal knowledge that tells an embodied agent both what must be true before an action and what changes as a result?

To this end, we introduce CausGround, a non-parametric causal interaction grounding framework for adaptive embodied planning. As illustrated in Fig.~\ref{fig:teaser}, CausGround abstracts interaction experience into action-centered causal knowledge, representing each action through its relevant preconditions and resulting state effects. Rather than storing complete trajectories, it organizes reusable action-centered causal knowledge over a factorized world-state representation that jointly captures object attributes and inter-object relations. 
At planning time, the current observation and task are grounded into the same representation to construct a task-relevant local causal graph from reusable causal knowledge. This allows the planner to reason backward over action preconditions to determine what must happen before an action becomes executable, and forward over state effects to anticipate how the world should change after execution. Composing such causal dependencies provides a structured reference for multi-step planning without updating model parameters.

We evaluate CausGround across heterogeneous long-horizon embodied settings. 
In simulation, we adapt VirtualHome~\citep{puig2018virtualhome} and RoboCasa~\citep{nasiriany2026robocasa365} for visually grounded high-level planning and observe consistent improvements under both one-shot and closed-loop planning, including an increase in average task success from 24.9\% to 53.3\% on VirtualHome and from 43.0\% to 57.0\% under closed-loop planning on RoboCasa.
Beyond simulation, we evaluate interaction reasoning on SWITCH~\citep{cheng2025switch}, a real-world egocentric long-horizon interaction benchmark, where CausGround improves average accuracy from 25.8\% to 31.9\%.
We further validate the framework on real-robot manipulation, where CausGround reasons backward from interpreted goal states to derive executable high-level action sequences and successfully complete multi-step interactions.
Together, these results show that explicit causal interaction knowledge provides a reusable grounding substrate for long-horizon embodied reasoning across diverse environments.

Our contributions are summarized as follows:
\begin{itemize}
    \item We formulate embodied interaction knowledge around action preconditions and state effects, and introduce a factorized world-state representation that jointly captures object attributes and inter-object relations, enabling interaction knowledge to be grounded in current visual observations.

    \item We introduce CausGround, a non-parametric framework that distills interaction trajectories into reusable action-centered causal knowledge and dynamically grounds task-relevant causal dependencies for planning. This enables reasoning over both what must hold before an action and how the world changes after it without parameter updates.

    \item We evaluate CausGround across long-horizon embodied settings spanning VirtualHome and RoboCasa simulations, the real-world egocentric SWITCH benchmark, and real-robot manipulation, demonstrating consistent gains in planning and interaction reasoning across environments and MLLM backbones.
\end{itemize}

%% file: Content/related.tex
\section{Related Works}
\label{sec:related-works}

\textbf{Environment Adaptation for Embodied Agents.}
Embodied agents deployed in new environments often need to adjust their behavior based on newly acquired interaction experience.
Earlier approaches typically absorb environment-specific experience into model parameters through fine-tuning, imitation learning, or reinforcement learning~\citep{kim2026robot, kang2025gflowvlm, tian2026seea}.
With the growing capabilities of foundation models, recent work has increasingly explored adaptation without continual parameter updates, using execution feedback~\citep{wang2023voyager}, past trajectories~\citep{zhao2024expel}, or external memory~\citep{sohn2026r4} to provide environment-specific information during deployment. 
This shifts environment adaptation from updating the model itself toward leveraging external environment knowledge. Therefore, effective adaptation depends not only on how experience is acquired and accumulated, but also on how the underlying world state is represented.

\textbf{World State Representations for Embodied Planning.}
Existing world-state representations vary substantially in their level of abstraction.
Free-form textual representations can flexibly summarize observations and state changes, but often entangle multiple entities and dynamic properties within a shared description~\citep{hao2023reasoning, chen2025planning}.
Object-centric representations reduce this entanglement by organizing state information around individual entities~\citep{yoneda2024statler, feng2026learning}, while more factorized representations further separate each object into explicit attribute-value pairs, enabling finer-grained state tracking~\citep{shen2026reward, rozanov2025stateact}. Yet embodied interaction depends not only on the state of individual objects, but also on relations among them~\citep{ju2026momagraph, gu2024conceptgraphs}. 
Therefore, a world-state representation for embodied planning should jointly capture fine-grained object attributes and explicit inter-object relations, providing a shared basis for grounding observations and environment-specific interaction knowledge.

\textbf{Interaction-Grounded High-Level Planning.}
A structured world state provides a basis for grounding environment-specific knowledge, but effective planning further requires turning such knowledge into actionable guidance.
MLLM-based planners can generate actions or action sequences from current observations and task descriptions~\citep{lv2024robomp, feng2025reflective}, leveraging strong pretrained priors to guide task completion. Before committing actions to the environment, world-model-based planners further predict action-conditioned state changes~\citep{li2026word, song2026model} and evaluate candidate trajectories through simulation or adaptive latent-space rollouts~\citep{liu2026simpact, yao2026dronewam, RISE}, providing lookahead for multi-step decisions.
However, while world models support reasoning about action consequences, action conditions are not always explicitly represented, which can make action proposal less reliable. Moreover, dependencies across multiple actions are typically instantiated only through rollout, requiring repeated simulation during planning. CausGround instead exposes action conditions, effects, and cross-action dependencies as explicit environment-specific knowledge that can be directly reused by the planner.

%% file: Content/methods.tex
\section{Methodology}
\label{sec:methods}

\begin{figure*}[t]
    \centering
    \includegraphics[width=0.90\textwidth]{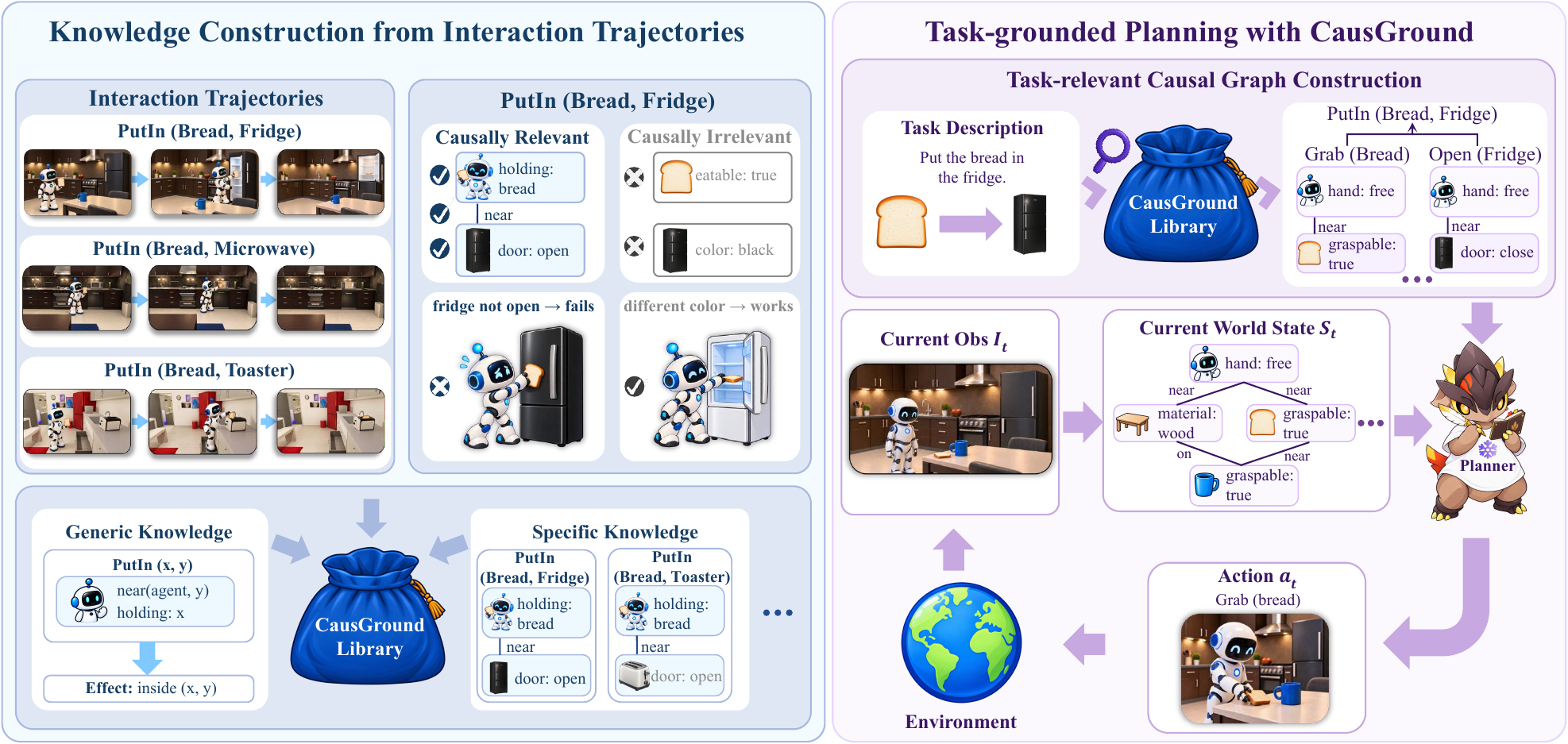}
    \caption{
    \textbf{Overview of CausGround.}
    \textbf{Left:} CausGround distills interaction trajectories into reusable action-centered causal knowledge by extracting action preconditions and effects, which are organized into generic and instance-specific knowledge.
    \textbf{Right:} Given the task description and current world state, task-relevant causal knowledge is grounded to construct a causal graph that provides executable guidance for a frozen planner.
    }
    \label{fig:method}
\end{figure*}

We aim to adapt embodied planners to environment-specific interaction dynamics without updating model parameters.
To this end, we introduce CausGround, which distills interaction experience into reusable action-centered causal knowledge that captures both action preconditions and state effects.
As illustrated in Fig.~\ref{fig:method}, CausGround consists of two main stages: causal interaction knowledge construction and task-grounded planning.
Specifically, we first define a factorized world-state representation that serves as a shared interface between current visual observations and causal interaction knowledge (Sec.~\ref{sec:factorized_state}). 
We then construct reusable action-centered causal knowledge over this representation by extracting action preconditions and effects from environment interactions (Sec.~\ref{sec:causal_construction}). 
Finally, we ground task-relevant causal knowledge in the current task and world state to support planning with a frozen planner (Sec.~\ref{sec:task_planning}).

\subsection{Factorized World-State Representation}
\label{sec:factorized_state}
To bridge causal knowledge construction and task-time grounding, CausGround represents the world state extracted from current observations and the preconditions and effects captured by causal knowledge using the same factorized schema (Fig.~\ref{fig:method}).
Given an RGB observation $I_t$, a frozen MLLM extracts the corresponding factorized world state $S_t$, exposing explicit state factors for matching the action preconditions and effects stored in CausGround. Specifically, we represent the world state at time $t$ as:
\begin{equation}
    S_t = (O_t, R_t),
\end{equation}
where $O_t$ is the set of objects in the scene and $R_t$ is the set of explicit inter-entity relations. 
Each object $o_i \in O_t$ is represented by its semantic category $c_i$ and a set of attribute-value pairs $A_i$,
\begin{equation}
    o_i = (c_i, A_i), \qquad
    A_i = \{(k_{ij}, v_{ij})\}_{j=1}^{M_i},
\end{equation}
where $k_{ij}$ and $v_{ij}$ denote the $j_{th}$ attribute key and value of object $o_i$, respectively, and $M_i$ is the number of its attributes. 
Each relation $r_{ij} \in R_t$ is represented as a typed edge between two objects (e.g.,~\texttt{ON(cup, table)},~\texttt{INSIDE(cup, fridge)}).
Each attribute-value pair and inter-object relation is treated as an explicit state factor that can serve as an action precondition or effect.
This representation explicitly separates object-local attributes from inter-object relations while preserving both within a unified world state.

Since observations and causal knowledge share the same factorized schema, action preconditions and effects can be directly matched against the current world state.
Meanwhile, the factorized representation makes individual state factors explicitly identifiable, allowing CausGround to identify individual factors involved in action execution and state transitions.

\subsection{Action-Centered Causal Knowledge Construction}
\label{sec:causal_construction}
Building on the factorized world state, CausGround organizes interaction experience at the action level (e.g.,~\texttt{GRAB},~\texttt{OPEN}), using actions as atomic units for extracting reusable preconditions and state effects.
For an action family $a$, we collect a set of interaction trajectories under controlled variations of candidate pre-action factors:
\begin{equation}
    \mathcal{T}_a = \{(S_i^{\mathrm{pre}}, a(x_i), S_i^{\mathrm{post}}, y_i)\}_{i=1}^{|\mathcal{T}_a|},
\end{equation}
where $S_i^{\mathrm{pre}}$ and $S_i^{\mathrm{post}}$ are the pre and post-action world states, $x_i$ denotes the object or arguments involved in the grounded action, and $y_i$ is the execution outcome.

CausGround abstracts action preconditions and effects from accumulated action-level interaction trajectories. Each pre-action state provides a set of candidate state factors. More specifically, each object attribute-value pair or inter-object relation is treated as an atomic candidate precondition.
For each grounded action $a(x)$, CausGround aggregates its corresponding trajectories in $\mathcal{T}_a$, identifies pre-action state factors whose satisfaction consistently distinguishes successful executions from failures, and retains them in the precondition set $C(a(x))$, while incidental factors are discarded.
For action effects, CausGround compares the corresponding pre and post-action states of successful executions. State changes that consistently follow action execution are retained in the effect set $E(a(x))$, while incidental changes are discarded.
Together, $C(a(x))$ and $E(a(x))$ characterize what must hold before the grounded action can be executed and what state changes follow from its execution. When an effect of one action satisfies a precondition of another, CausGround connects the two actions through a causal dependency, allowing multi-step interactions to be composed through such dependencies.

Moreover, CausGround organizes the extracted precondition and effect sets according to their level of specificity.
Across grounded instances of the same action family $a$ (e.g.,~\texttt{OPEN(fridge)} and~\texttt{OPEN(microwave)}), CausGround compares their precondition and effect sets to identify causal regularities that generalize across instances. Preconditions and effects that are consistently shared across multiple instances are abstracted into generic action knowledge $K_{\mathrm{gen}}(a)$, while regularities that remain valid only for particular objects are preserved as instance-specific knowledge $K_{\mathrm{spec}}(a(x))$. For example,~\texttt{OPEN(fridge)} and~\texttt{OPEN(microwave)} may share the effect that~\texttt{openness} becomes~\texttt{open}, while requiring different object-specific preconditions for successful execution. In this way, generic knowledge captures reusable causal regularities across grounded instances, while instance-specific knowledge preserves valid object-specific variations. As shown in the left part of Fig.~\ref{fig:method}, these two levels together form the reusable action-centered causal knowledge used for task-time grounding.

\subsection{Task-Grounded Planning}
\label{sec:task_planning}
As illustrated in the right part of Fig.~\ref{fig:method}, CausGround grounds reusable causal knowledge with the current task and world state before providing it to the frozen planner.
Specifically, given the initial world state $S_0$ and task description $g$, CausGround constructs a task-relevant causal graph through two complementary paths.
First, we interpret $g$ into a goal state under the same factorized world-state representation and perform backward reasoning from the goal state over action effects and preconditions, yielding goal-directed causal dependencies $K_{\mathrm{causal}}$.
In parallel, we semantically retrieve related generic and instance-specific causal knowledge from the CausGround library, denoted as $K_{\mathrm{sem}}$. The final task knowledge is:
\begin{equation}
    K_{\mathrm{task}} = K_{\mathrm{causal}} \cup K_{\mathrm{sem}}.
\end{equation}

$K_{\mathrm{task}}$ is constructed once at the beginning of each task and grounded with the current world state, allowing the frozen planner to reason over task-relevant causal dependencies. At each step, the planner combines the current visual observation and its structured world state with the task description and retrieved task knowledge to select the next action:
\begin{equation}
    a_t = \pi_{\theta}(I_t, S_t, g, K_{\mathrm{task}}).
\end{equation}
The grounded causal knowledge provides bidirectional guidance for planning: backward reasoning over action preconditions identifies executable actions, forward reasoning over action effects supports state-transition reasoning, and cross-action dependencies provide structured guidance for long-horizon planning.

%% file: Content/experiments.tex
\section{Experiments}
\label{sec:experiments}

\begin{figure*}[t]
    \centering
    \includegraphics[width=0.88\textwidth]{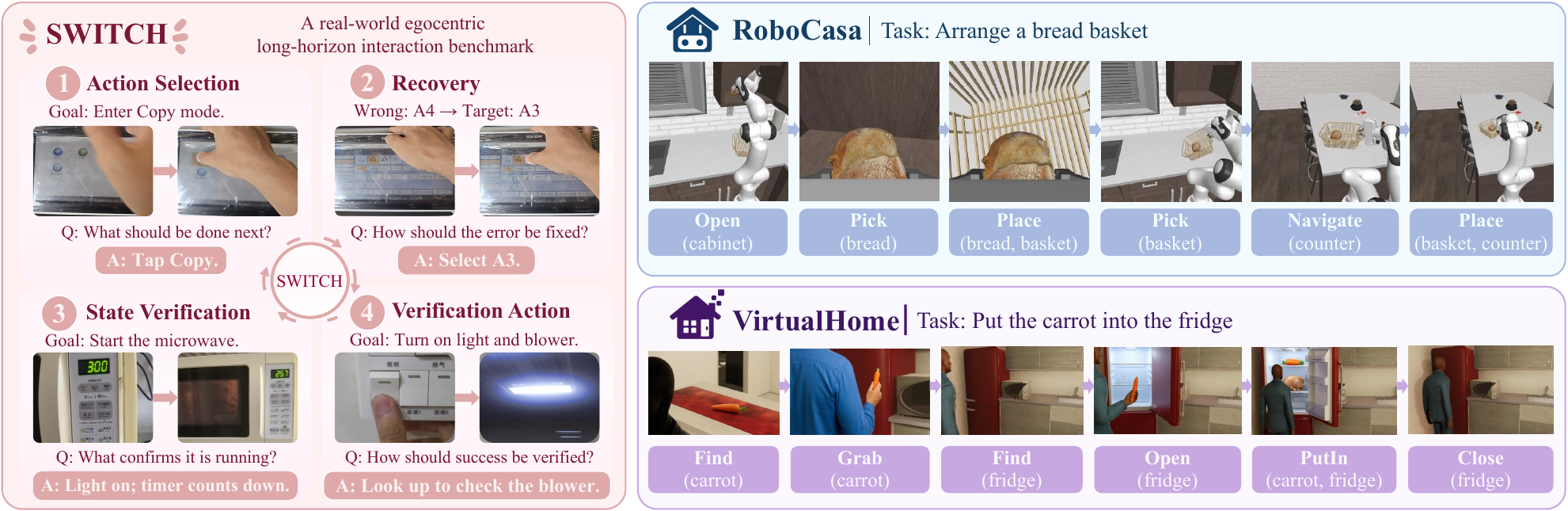}
    \caption{
    Evaluation environments and representative tasks.
    \textbf{Left:} SWITCH examples covering action selection, error recovery, state verification, and verification action.
    \textbf{Right:} High-level planning trajectories in RoboCasa and VirtualHome, showing post-action observations. RoboCasa steps correspond to annotated skill stages; VirtualHome uses repaired image observations.
    }
    \label{fig:eval_environments}
\end{figure*}

\subsection{Experimental Setup}
\textbf{Benchmarks and Evaluation Protocols.}
We evaluate CausGround from two complementary perspectives: visual high-level planning and interaction reasoning. As illustrated in Fig.~\ref{fig:eval_environments}, we use VirtualHome~\citep{puig2018virtualhome} and RoboCasa~\citep{nasiriany2024robocasa, nasiriany2026robocasa365} to evaluate long-horizon planning from visual observations, and SWITCH~\citep{cheng2025switch} to assess fine-grained reasoning over states, actions, and interaction outcomes.
Beyond benchmark evaluation, we further validate CausGround on real-world sequential manipulation with a physical robot.

For~\textbf{VirtualHome}, a simulated household environment with diverse multi-step activities, we focus on the image-based planning setting rather than its text-based counterpart. We correct rendered observations that are inconsistent with the underlying environment state using an image-editing model, with details provided in Appendix~\ref{app:virtualhome}. We report success rate (SR) and task progress, where progress measures partial completion of the task-specific semantic objectives at episode termination. To examine performance across planning horizons, we further partition tasks into short, medium, and long-horizon subsets according to the number of ground-truth high-level actions.

For~\textbf{RoboCasa}, a simulation benchmark for household manipulation, we adapt its tasks to high-level visual planning by treating the officially annotated skill stages as atomic actions and using the corresponding simulator observations as intermediate states. We report task success rate under both one-shot planning and closed-loop replanning.

For~\textbf{SWITCH}, which evaluates a broader range of embodied interaction reasoning capabilities, we focus on four categories that most directly probe action-level state-transition reasoning:~\textit{Action},~\textit{Recovery},~\textit{Verification State}, and~\textit{Verification Action}. Together, these categories cover action selection, error recovery, outcome verification, and active verification, forming a complementary reasoning loop around action applicability and effects. We report prediction accuracy for each category.

\textbf{Implementation Details.}
For each comparison, all methods are evaluated under the same planner backbone, visual observations, task descriptions, and high-level action space. In our main experiments, we instantiate the planner with Qwen3.5-9B, Qwen3.5-27B, and Qwen3.5-35B-A3B, while additional model families are considered in Sec.~\ref{sec:cross_family}. We compare CausGround against a~\textbf{Vanilla} planner and two latest methods, \textbf{SelfReVision}~\citep{park2025making} and \textbf{LookPlanGraph}~\citep{onishchenko2025lookplangraph}.
SelfReVision iteratively critiques, revises, and verifies generated procedural plans, while LookPlanGraph maintains a dynamically updated scene-memory graph from visual observations to ground planning.
We adapt both methods to the same visual observations and high-level action space used by CausGround.
All planner parameters remain frozen throughout evaluation. Additional implementation details, prompts, and baseline adaptations are provided in Appendix~\ref{app:implementation}.

\textbf{Overview.}
Our experiments progressively examine CausGround from effectiveness to generalization and analysis. We first evaluate whether CausGround improves visual high-level planning across different task horizons on VirtualHome and RoboCasa (Sec.~\ref{sec:visual_planning}). Apart from the vanilla planning performance, we evaluate interaction reasoning on SWITCH (Sec.~\ref{sec:switch}) and generalization across different model families and scales (Sec.~\ref{sec:cross_family}). Next, we isolate the contributions of the local world-state graph and causal interaction knowledge through component analysis (Sec.~\ref{sec:ablation}). Finally, we qualitatively analyze how CausGround changes planning behavior in VirtualHome and investigate whether simulation-acquired interaction knowledge can support real-world robot planning (Sec.~\ref{sec:qualitative}).

\begin{table*}[t]
\centering
\caption{
\textbf{High-level planning performance.}
VirtualHome reports success rate (SR) and final progress (P) across short, medium, and long horizon tasks, with Average denoting the macro-average over the three difficulty levels.
RoboCasa reports task success rate under one-shot planning and closed-loop replanning.
Best results are shown in~\textbf{bold} and second-best results are~\underline{underlined}.
}
\label{tab:main_results}

\renewcommand{\arraystretch}{1.10}

\resizebox{\textwidth}{!}{%
\begin{tabular}{llcccccccccc}
\toprule
\multirow{3}{*}{Backbone}
& \multirow{3}{*}{Method}
& \multicolumn{8}{c}{VirtualHome}
& \multicolumn{2}{c}{RoboCasa} \\
\cmidrule(lr){3-10}
\cmidrule(lr){11-12}

&
& \multicolumn{2}{c}{Short}
& \multicolumn{2}{c}{Medium}
& \multicolumn{2}{c}{Long}
& \multicolumn{2}{c}{Average}
& \multirow{2}{*}{One-shot}
& \multirow{2}{*}{Replan} \\

\cmidrule(lr){3-4}
\cmidrule(lr){5-6}
\cmidrule(lr){7-8}
\cmidrule(lr){9-10}

&
& SR & P
& SR & P
& SR & P
& SR & P
& & \\
\midrule

\multirow{4}{*}{Qwen3.5-9B}
& Vanilla
& 28.0 & 34.6
& 4.0 & 18.1
& 0.0 & 18.9
& 10.7 & 23.8
& \underline{31.0} & \underline{36.7} \\

& SelfReVision
& \underline{48.0} & 52.7
& \underline{12.0} & 22.5
& \underline{8.0} & \underline{27.2}
& \underline{22.7} & 34.1
& 26.8 & 26.5 \\

& LookPlanGraph
& 44.0 & \underline{56.8}
& \underline{12.0} & \underline{35.5}
& 4.0 & 26.9
& 20.0 & \underline{39.7}
& 19.7 & 28.6 \\

\rowcolor{oursblue}
& \textbf{CausGround}
& \textbf{80.0} & \textbf{83.6}
& \textbf{32.0} & \textbf{51.3}
& \textbf{20.0} & \textbf{49.8}
& \textbf{44.0} & \textbf{61.5}
& \textbf{45.1} & \textbf{46.9} \\

\midrule

\multirow{4}{*}{Qwen3.5-27B}
& Vanilla
& 60.0 & 63.3
& 28.0 & 45.6
& 12.0 & 31.9
& 33.3 & 46.9
& \textbf{27.9} & \underline{57.8} \\

& SelfReVision
& \underline{68.0} & \underline{69.3}
& 36.0 & 51.8
& 24.0 & 43.8
& 42.7 & 55.0
& 18.3 & 39.1 \\

& LookPlanGraph
& \textbf{84.0} & \textbf{84.0}
& \underline{52.0} & \underline{67.6}
& \underline{32.0} & \underline{63.3}
& \underline{56.0} & \underline{71.6}
& \underline{24.4} & 53.1 \\

\rowcolor{oursblue}
& \textbf{CausGround}
& \textbf{84.0} & \textbf{84.0}
& \textbf{60.0} & \textbf{78.3}
& \textbf{44.0} & \textbf{77.2}
& \textbf{62.7} & \textbf{79.8}
& \textbf{27.9} & \textbf{75.0} \\

\midrule

\multirow{4}{*}{Qwen3.5-35B-A3B}
& Vanilla
& 56.0 & 59.4
& 16.0 & 30.4
& \underline{20.0} & \underline{43.5}
& 30.7 & 44.4
& \underline{21.9} & \underline{34.4} \\

& SelfReVision
& 52.0 & 55.3
& \underline{40.0} & \underline{52.7}
& \underline{20.0} & 35.8
& \underline{37.3} & \underline{47.9}
& 15.1 & 4.9 \\

& LookPlanGraph
& \underline{64.0} & \underline{67.3}
& 28.0 & 45.5
& 8.0 & 28.9
& 33.3 & 47.2
& 19.2 & 23.0 \\

\rowcolor{oursblue}
& \textbf{CausGround}
& \textbf{88.0} & \textbf{90.0}
& \textbf{44.0} & \textbf{68.1}
& \textbf{28.0} & \textbf{59.5}
& \textbf{53.3} & \textbf{72.6}
& \textbf{43.8} & \textbf{49.2} \\

\bottomrule
\end{tabular}%
}

\end{table*}

\subsection{High-Level Planning Performance}
\label{sec:visual_planning}
As shown in Table~\ref{tab:main_results}, CausGround achieves the best performance across all planner backbones and task difficulties on VirtualHome, improving the average success rate from 24.9\% with Vanilla to 53.3\% and the average final progress from 38.4\% to 71.3\% over Vanilla.
Although planning becomes increasingly challenging as the horizon grows, CausGround consistently maintains both higher success rates and stronger task progress from short to long-horizon tasks.
This suggests that causal interaction knowledge not only helps complete more tasks, but also enables the planner to make more sustained progress over extended interaction sequences.

CausGround shows the same overall advantage on RoboCasa, achieving the strongest performance in most backbones and protocol settings, and improving one-shot planning and closed-loop replanning. Averaged across backbones, CausGround improves one-shot success from 26.9\% to 38.9\% and replanning success from 43.0\% to 57.0\% over Vanilla.
Interestingly, adding more reasoning or structure does not necessarily yield better planning: SelfReVision and LookPlanGraph can even underperform Vanilla. We find that free-form self-reflection may introduce unsupported prerequisites and consequently corrupt an originally valid plan (e.g., incorrectly infer that an object is inside a cabinet and insert unnecessary~\texttt{OPEN} actions). Similarly, discrete scene graphs used in LookPlanGraph can propagate grounding errors into subsequent decisions (e.g., misinterpreting a closed microwave as open), causing the planner to skip necessary actions and produce an invalid plan.

Above all, the benefit of CausGround is not simply to provide more reasoning context, but to supply a task-relevant causal subgraph of action preconditions and effects that is grounded against the current observation before each planning decision.
The grounded knowledge supports state-aware action proposal and structured reference for multi-step planning, helping the planner maintain long-horizon progress while adapting to the current executable state and avoiding repeated failures or deadlocks.

\subsection{Causal State-Transition Reasoning on SWITCH}
\label{sec:switch}
Beyond end-to-end planning, we evaluate interaction reasoning on SWITCH~\citep{cheng2025switch}, which measures an agent's ability to reason over action execution and state transitions under long-horizon interactions.
As shown in Table~\ref{tab:switch_results}, CausGround consistently improves Vanilla across all MLLM backbones, increasing average accuracy from 25.8\% to 31.9\%.
The improvements are consistent across different aspects of interaction reasoning (e.g.,~\textit{Action},~\textit{Recovery},~\textit{Verification Action}, and~\textit{Verification State}), demonstrating that causal interaction knowledge improves reasoning over both action selection and resulting state transitions.

\begin{table}[t]
\centering
\caption{
\textbf{Interaction reasoning on SWITCH.}
Accuracy (\%) is reported across four interaction reasoning categories and three different backbones. Best results are shown in bold.
}
\label{tab:switch_results}

\small
\setlength{\tabcolsep}{3.5pt}
\renewcommand{\arraystretch}{1.00}

\begin{tabular}{@{}llccccc@{}}
\toprule
Backbone
& Method
& Action
& Recovery
& Verification Action
& Verification State
& Average
\\
\midrule

\multirow{2}{*}{Qwen3.5-9B}
& Vanilla
& 24.2
& 36.8
& 30.2
& 19.6
& 27.7
\\
& \ourscell{\textbf{CausGround}}
& \ourscell{\textbf{28.1}}
& \ourscell{\textbf{44.7}}
& \ourscell{\textbf{36.5}}
& \ourscell{\textbf{22.2}}
& \ourscell{\textbf{32.9}}
\\

\midrule

\multirow{2}{*}{Qwen3.5-27B}
& Vanilla
& 31.9
& 28.9
& 36.5
& 14.2
& 27.9
\\
& \ourscell{\textbf{CausGround}}
& \ourscell{\textbf{33.7}}
& \ourscell{\textbf{39.5}}
& \ourscell{\textbf{52.4}}
& \ourscell{\textbf{15.1}}
& \ourscell{\textbf{35.2}}
\\

\midrule

\multirow{2}{*}{Qwen3.5-35B-A3B}
& Vanilla
& 27.1
& 23.7
& 23.8
& 12.0
& 21.7
\\
& \ourscell{\textbf{CausGround}}
& \ourscell{\textbf{27.4}}
& \ourscell{\textbf{28.9}}
& \ourscell{\textbf{41.3}}
& \ourscell{\textbf{13.3}}
& \ourscell{\textbf{27.7}}
\\

\bottomrule
\end{tabular}
\end{table}

\begin{figure}[t]
    \centering
    \includegraphics[width=0.90\textwidth]{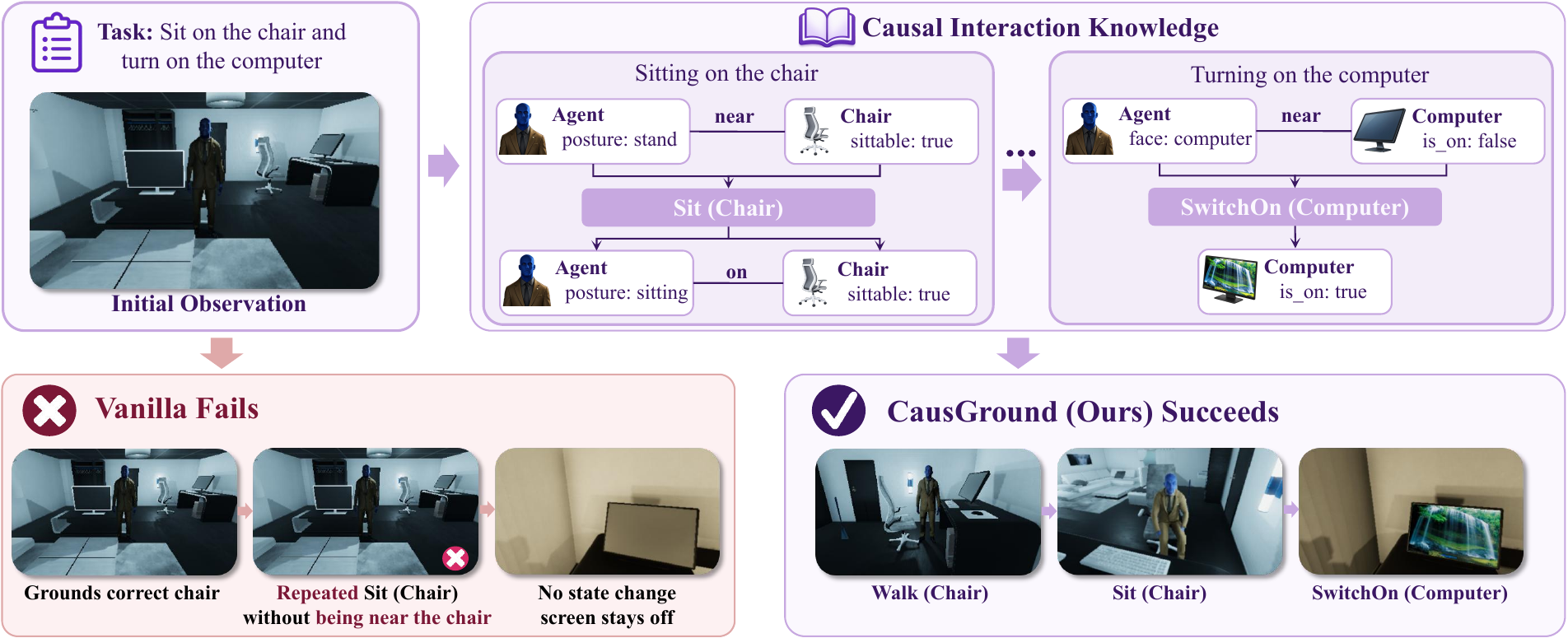}
    \caption{
    \textbf{Case Study.}
    Vanilla identifies the correct chair but repeatedly executes SIT without satisfying the NEAR precondition, while CausGround uses causal knowledge to complete the task.
    }
    \label{fig:case_study}
\end{figure}

\subsection{Component Ablations}
\label{sec:ablation}
We disentangle the contributions of the two components in CausGround.
Causal interaction knowledge provides reusable action preconditions and effects, while the factorized state explicitly organizes the task-relevant state.
As shown in Table~\ref{tab:ablation}, both components improve over Vanilla, but they contribute in different ways.
First, explicitly representing the task-relevant world state helps the agent identify what is true in the environment and make more reliable decisions.
\begin{wraptable}[10]{r}{0.40\textwidth}
    \vspace{-0.75\baselineskip}
    \centering

    \caption{
    \textbf{Component ablations.}
    Contributions of the local graph and causal interaction knowledge.
    }
    \label{tab:ablation}

    \small
    \setlength{\tabcolsep}{6pt}
    \renewcommand{\arraystretch}{1.18}

    \begin{tabular}{lcc}
    \toprule
    Method & SR & Progress \\
    \midrule
    Vanilla
    & 33.3 & 46.9 \\

    + Factorized State
    & 36.0 & 55.1 \\

    + Causal Knowledge
    & 53.3 & 70.8 \\

    \rowcolor{oursblue}
    \textbf{CausGround}
    & \textbf{62.7} & \textbf{79.8} \\
    \bottomrule
    \end{tabular}

\end{wraptable}
Second, causal interaction knowledge provides a structured view of when an action is applicable and how it changes the state, allowing the agent to anticipate consequences before execution.
Naturally, the full CausGround performs best overall.
Since the local state and causal knowledge share the same factorized state representation, the planner can directly match the current state against the action preconditions and effects described in the retrieved knowledge.
This grounds the causal knowledge to what is actually observed, making it useful for planning.

\subsection{Cross-Family Generalization}
\label{sec:cross_family}

\begin{wrapfigure}[15]{r}{0.44\textwidth}
    \vspace{-0.50\baselineskip}
    \centering
    \includegraphics[width=\linewidth]{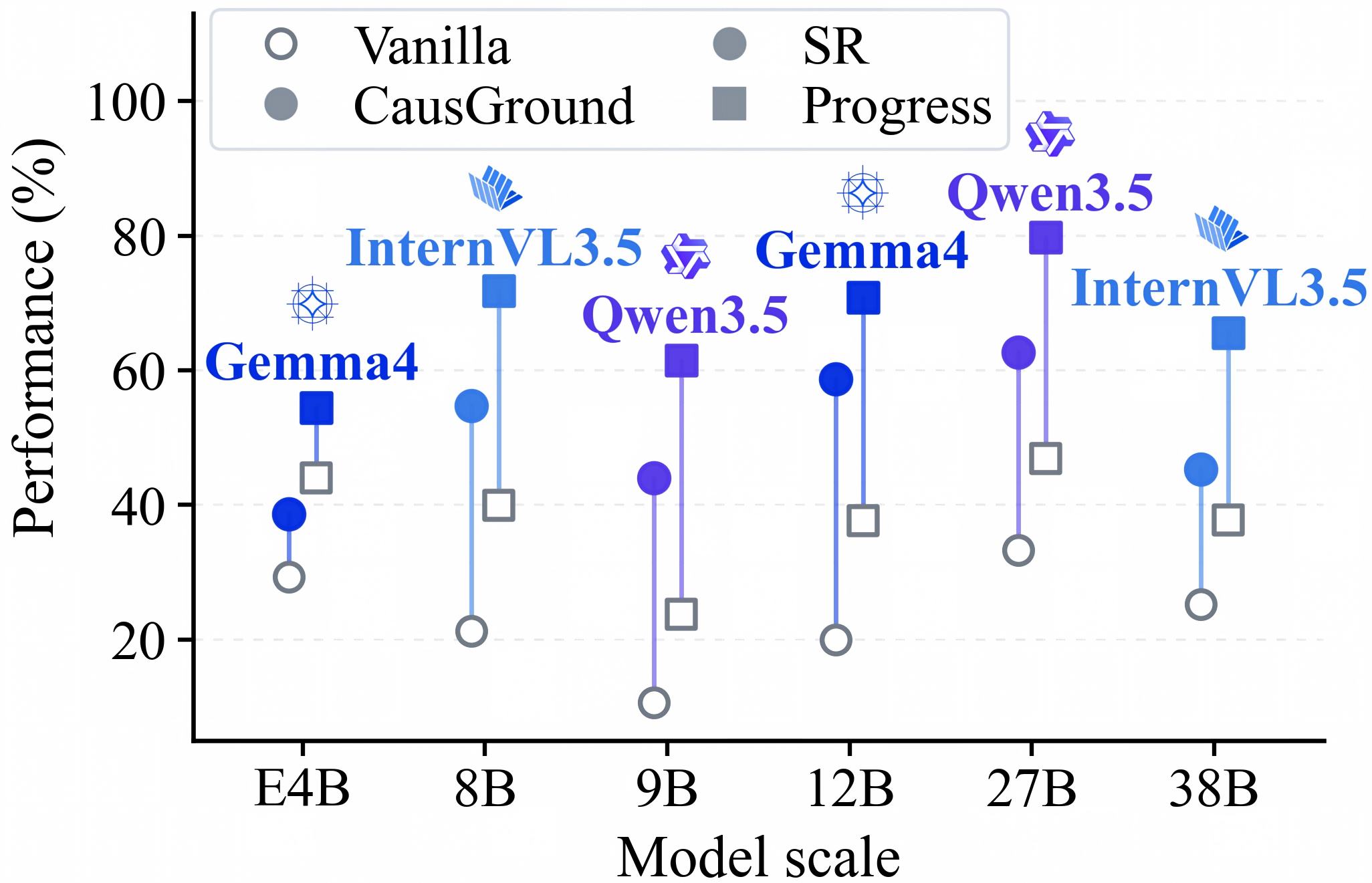}
    \vspace{-0.30\baselineskip}
    \caption{
    \textbf{Generalization across model families and scales.}
    CausGround improves SR and final progress across backbones.
    }
    \label{fig:model_generalization}
    \vspace{-0.35\baselineskip}
\end{wrapfigure}

To examine whether the benefit of CausGround depends on a particular planner backbone, we further evaluate six models from three families, including Qwen3.5, InternVL3.5, and Gemma4, covering model sizes from E4B to 38B.
As shown in Fig.~\ref{fig:model_generalization}, CausGround consistently improves both task success and final progress over Vanilla across all six backbones.
Averaged across these models, SR increases from 23.3\% to 50.7\%, while final progress improves from 38.4\% to 67.3\%.
Notably, the improvement is observed for both smaller and larger models and across all three model families, indicating that the benefit of CausGround is not tied to a particular model family or scale.
These results further support the use of reusable interaction knowledge as a general external prior for different planners.

\subsection{Qualitative Analysis}
\label{sec:qualitative}

\textbf{Case Study.}
Figure~\ref{fig:case_study} illustrates a representative failure mode of Vanilla. Although it identifies the target chair, Vanilla repeatedly executes SIT without satisfying the NEAR precondition. CausGround identifies the missing prerequisite through causal knowledge and completes the task.

\begin{figure*}[t]
    \centering
    \includegraphics[width=0.90\textwidth]{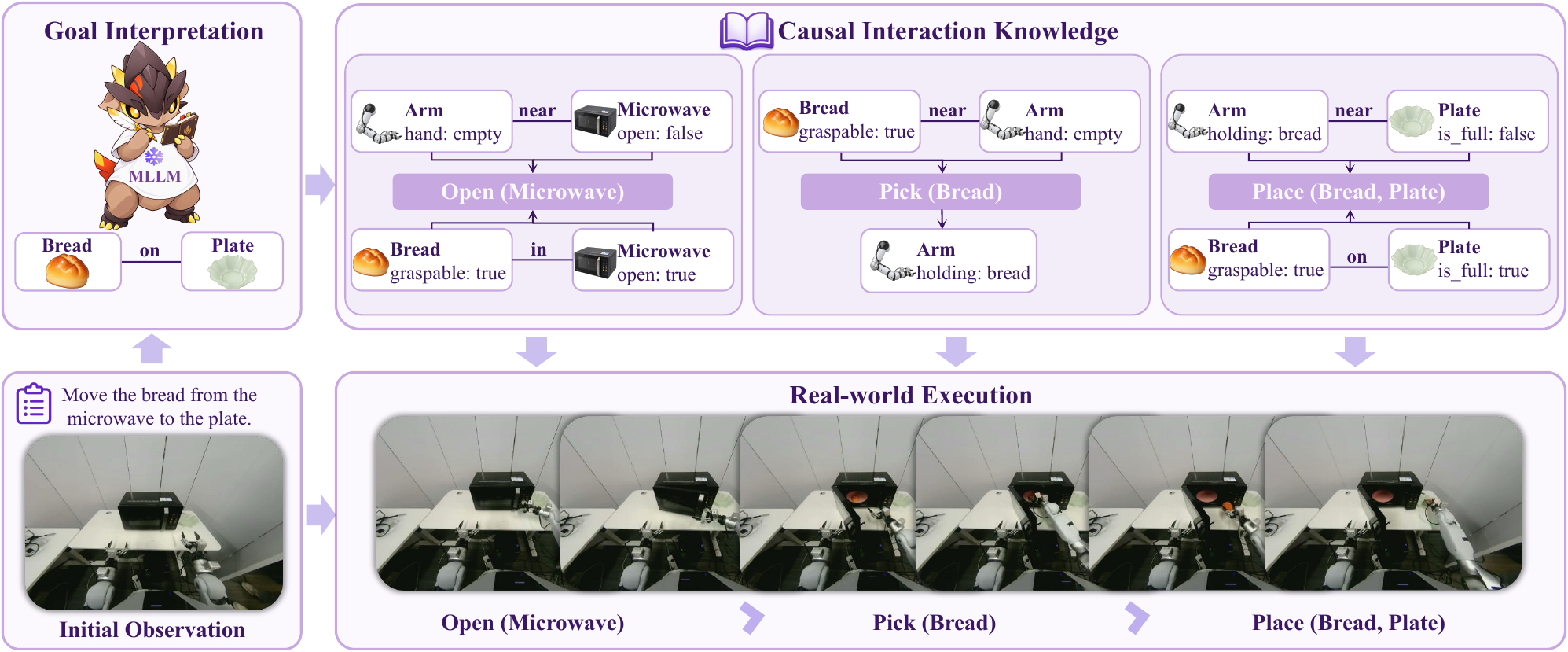}
    \caption{
    \textbf{Real-world Deployment.}
    Given initial observation and task, CausGround retrieves task-relevant causal knowledge from the interpreted goal state and grounds it to executable actions.
    }
    \label{fig:real_world}
\end{figure*}

\textbf{Real-World Deployment.}
We validate CausGround on real-world manipulation. Given the task and initial observation, the MLLM interprets the goal state, from which CausGround performs backward reasoning over causal dependencies to retrieve task-relevant causal knowledge and derive executable actions (Open(Microwave), Pick(Bread), and Place(Bread, Plate)), as shown in Fig.~\ref{fig:real_world}.

%% file: Content/conclusion.tex
\section{Conclusion}
\label{sec:conclusion}
\looseness=-1
We introduce CausGround, a causal interaction grounding framework for adapting MLLM-based embodied planners to environment-specific interaction dynamics without parameter updates.
Instead of storing interaction experience as unstructured trajectories, CausGround distills reusable action-centered causal knowledge by identifying action preconditions and effects, expressed under a shared factorized world-state schema that jointly captures object attributes and inter-object relations.
At planning time, CausGround grounds task-relevant causal knowledge with the current task and world state, supporting backward reasoning over action preconditions, forward reasoning over action effects, and long-horizon planning.
These results highlight explicit causal interaction knowledge as a reusable grounding mechanism for long-horizon embodied planning across diverse environments.

%% file: Content/appendix.tex
\newpage
\appendix

\section{Evaluation Environment Construction}
\label{app:environment}

\subsection{VirtualHome}
\label{app:virtualhome}

\subsubsection{Benchmark Construction and Task Statistics}
\label{app:virtualhome_benchmark}
We adapt VirtualHome, which has been predominantly studied through textual or symbolic task--action interfaces in prior LLM-based planning work~\citep{huang2022language}, into a \textbf{vision-based benchmark for closed-loop high-level planning}.
We construct a fixed suite of 75 household tasks and divide them into three difficulty levels according to the number of executable actions in their reference programs.
Specifically, Short tasks contain 3--6 reference actions, Medium tasks contain 7--10, and Long tasks contain 11--22.
Across all tasks, the benchmark spans six VirtualHome scenes, 33 distinct action types, and 78 distinct object classes, covering navigation, object manipulation, container and device interaction, posture changes, cleaning, and food-related activities.
Each split contains 25 tasks, yielding 643 reference actions in total.
Table~\ref{tab:virtualhome_statistics} summarizes the benchmark statistics.

\begin{table}[H]
\centering
\caption{\textbf{Statistics of the adapted VirtualHome benchmark.}
Tasks are divided into Short, Medium, and Long splits according to the number of reference actions per task.
The maximum action attempts specify the independent interaction budget available to the agent during closed-loop evaluation.}
\label{tab:virtualhome_statistics}
\setlength{\tabcolsep}{5.0pt}
\begin{tabular}{lcccc}
\toprule
& Short & Medium & Long & Total \\
\midrule
\# Tasks                       & 25    & 25     & 25      & 75 \\
Reference actions per task    & 3--6  & 7--10  & 11--22  & 3--22 \\
Mean reference actions per task
                               & 4.28  & 8.12   & 13.32   & 8.57 \\
Max. action attempts          & 20    & 40     & 60      & -- \\
\bottomrule
\end{tabular}
\end{table}

\subsubsection{Visual Planning Interface and Execution Protocol}
\label{app:virtualhome_interface}
We evaluate all methods in a closed-loop setting: at each step, the planner observes the current RGB scene, predicts one high-level action, the action is executed in VirtualHome, and the resulting observation is returned for the next decision.
To make this setting well-defined and comparable across methods, we use a shared planner interface specifying the observation, action space, object vocabulary, and execution feedback.

\setlength{\LTcapwidth}{\textwidth}

\begin{longtable}{p{0.25\linewidth}p{0.68\linewidth}}
\caption{\textbf{Shared planning interface in VirtualHome.}
All methods receive the same public task information, visual observations, action templates, object vocabulary, and execution feedback. Private simulator states and reference programs are not exposed to the planner.}
\label{tab:virtualhome_interface} \\

\toprule
Component & Definition \\
\midrule
\endfirsthead

\toprule
Component & Definition \\
\midrule
\endhead

\endfoot

\bottomrule
\endlastfoot

Task input
& A fixed natural-language task instruction. \\

Visual input
& Three simultaneous RGB observations rendered from the current simulator state. \\

Action history
& The planner's own previously attempted actions together with their SUCCESS/FAILURE outcomes. \\

Action space
& A fixed set of 45 atomic high-level action templates supported by the VirtualHome semantic executor, e.g., \texttt{WALK()}, \texttt{GRAB()}, \texttt{OPEN()}, \texttt{PUTIN()}, and \texttt{SWITCHON()}. \\

Object vocabulary
& A public menu of executable object-class names used to bind predicted action arguments to simulator entities; private simulator locations and hidden state are excluded. \\

Planner output
& One atomic action together with its object-class argument(s). \\

Execution feedback
& Whether the attempted action succeeds or fails; private executor failure reasons are not exposed. \\

Private information
& Full simulator state graphs, reference actions, task objectives, and GT action ordering are never provided to the planner. \\
\end{longtable}

The frozen 75-task benchmark exercises 33 distinct action types
(e.g., \texttt{OPEN(X)} and \texttt{PUTIN(X,Y)}), comprising 643 reference actions in total.
Action arguments are restricted to a public vocabulary of 308 canonical object classes
(e.g., \texttt{bread} and \texttt{microwave}) and are subsequently resolved to executable simulator entities, yielding actions such as \texttt{OPEN(microwave)} and \texttt{PUTIN(bread,microwave)}.

\subsubsection{Observation Repair}
\label{app:virtualhome_repair}
Although VirtualHome maintains an explicit simulator state for interaction execution, its native RGB renderer can occasionally produce observations that are visually inconsistent with the underlying state after certain interactions.
Such inconsistencies are particularly problematic in our vision-based setting, since the planner must infer the current world state from RGB observations rather than privileged symbolic information.
We therefore introduce an observation-repair procedure that corrects state-inconsistent visual content while preserving the original camera viewpoint, scene layout, object identities, and unrelated scene content.

For each inconsistent observation, the repair procedure takes the native rendered frame together with the executed action and the corresponding simulator-derived state transition as input.
We use Qwen-Image-Edit-2511~\citep{wu2025qwen} to generate a corrected observation under a constrained editing prompt.
The edit is restricted to visual changes implied by the executed action, while the camera viewpoint, scene geometry, object identities, appearance, and unrelated scene content are preserved.
The repair is not allowed to introduce future states, hallucinate occluded or out-of-view objects, or modify objects unrelated to the current transition.

Each repaired observation is checked against the resulting simulator state using deterministic consistency rules and manual visual inspection.
Only validated observations are retained for evaluation.
The privileged simulator state used during this repair and validation process is never exposed to the planner and serves solely to ensure the RGB observation is consistent with the actual environment.

\textbf{Repair Prompt.}
For reproducibility, we summarize the core prompt used for observation repair.

\begingroup
\setlength{\parskip}{0pt}
\tcbset{paperprompt/.append style={breakable}}

\begin{tcolorbox}[
    paperprompt,
    width=\textwidth,
    breakable,
    coltitle=black,
    fontupper=\fontsize{9}{9.5}\selectfont,
    fonttitle=\fontsize{9}{10}\selectfont,
    left=6pt,
    right=6pt,
    top=3pt,
    bottom=3pt,
    boxsep=0pt,
    toptitle=2pt,
    bottomtitle=2pt,
    before skip=4pt,
    after skip=4pt,
    colback=white,
    colframe=promptpinkframe,
    colbacktitle=promptpinktitle,
    title={
        \textbf{Observation-Repair Prompt}
        \hfill
        \textit{Qwen-Image-Edit-2511}
    },
    title after break={
        \textbf{Observation-Repair Prompt}
        \hfill
        \textit{Qwen-Image-Edit-2511}
    }
]

\textbf{Native Observation:}\par
\texttt{\{rendered\_observation\}}\par

\vspace{2pt}
\textbf{Executed Action:}\par
\texttt{\{action\}}\par

\vspace{2pt}
\textbf{State Transition:}\par
\texttt{\{pre\_state\} $\rightarrow$ \{post\_state\}}\par

\vspace{2pt}
\textbf{Relevant Entity Binding:}\par
\texttt{\{actor / manipulated objects / containers / destinations\}}\par

\vspace{2pt}
\begin{enumerate}[
    leftmargin=*,
    itemsep=0.5pt,
    topsep=2pt,
    parsep=0pt,
    partopsep=0pt
]

\item \textbf{Objective.}
Edit the native observation only when necessary so that the visible scene is consistent with the actual post-action simulator state.
The executed action and state transition specify what has changed; they must not be interpreted as instructions to modify unrelated content.

\item \textbf{Apply only the current state change.}
Modify only visual content directly implied by the executed action and its resulting state transition.
Do not introduce effects from future actions, previous unexecuted actions, or hypothetical task progress.

\item \textbf{Preserve unaffected scene content.}
Keep the camera viewpoint, framing, scene layout, character identity, object identities, geometry, materials, scale, lighting, and all unrelated objects unchanged.

\item \textbf{Maintain interaction consistency.}
The edited image must reflect the resulting interaction state.
For example, after a successful \texttt{GRAB(X)}, the object should be held by the actor and should no longer remain at its source location;
after a successful placement action, the object should appear at the specified destination and should no longer be held.

\item \textbf{Respect visibility.}
Do not hallucinate objects that are occluded, inside a closed container, or outside the current camera view.
Do not expose hidden simulator information merely because it is available in the symbolic state.

\item \textbf{Avoid unrelated edits.}
Do not add, remove, replace, recolor, relocate, resize, or alter any entity unrelated to the current state transition.
Do not change the task layout, camera pose, character appearance, or background scene.

\item \textbf{Output requirement.}
Return one repaired RGB observation that is visually consistent with the post-action simulator state while remaining maximally faithful to the native render.

\end{enumerate}

\end{tcolorbox}
\endgroup

\begin{figure}[H]
    \centering
    \includegraphics[width=\textwidth]{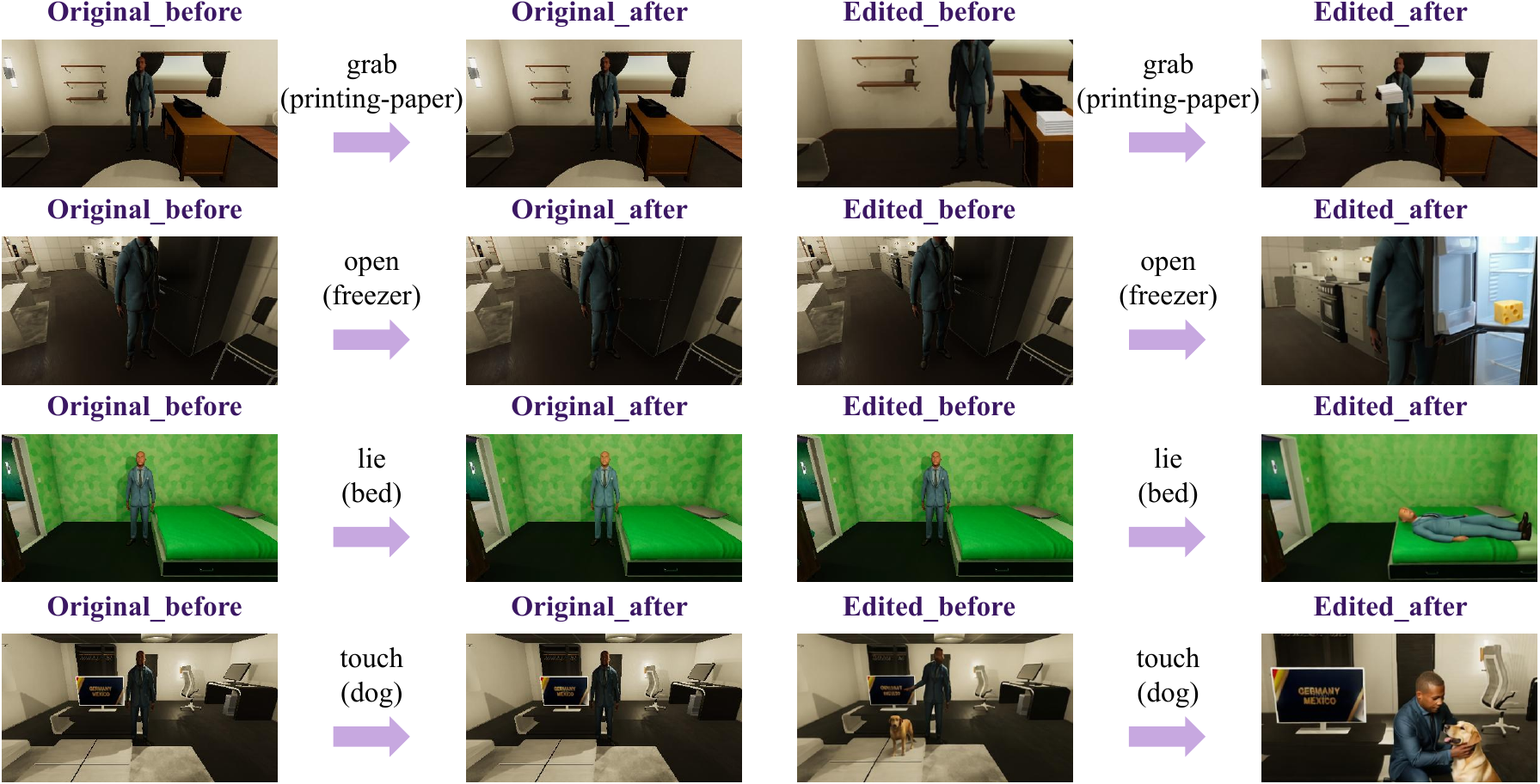}
    \caption{\textbf{Representative observation repairs in VirtualHome.}
    Each row shows a transition example with the original pre and post observations (left two columns) and the repaired pre/post observations (right two columns).
    The repair corrects visually inconsistent state transitions, such as object transfer, container interaction, and posture change, while preserving unrelated scene content.}
    \label{fig:virtualhome_repair}
\end{figure}

\textbf{Repair Examples.}
Figure~\ref{fig:virtualhome_repair} shows representative before-and-after observation repairs in VirtualHome.
The repair modifies only state-inconsistent visual content while preserving the original viewpoint, scene layout, and unrelated objects.

\subsubsection{Evaluation Protocol}
\label{app:virtualhome_evaluation}

Each task is associated with a private semantic completion specification defined by its target outcome and required interaction events, with ordering constraints included when necessary.
We report \textbf{Success Rate (SR)} as the percentage of episodes that satisfy the complete task specification within the interaction budget.
\textbf{Progress (P)} measures partial task completion according to the same semantic specification at the end of an episode.

Importantly, evaluation does not require the planner to reproduce the reference action sequence.
The reference programs are used only for benchmark construction, difficulty stratification, and private evaluation, allowing alternative valid action sequences to receive full credit when they achieve the required task outcome.
An episode terminates once the task is completed or the maximum number of action attempts is reached.
The interaction budgets are 20, 40, and 60 attempts for Short, Medium, and Long tasks, respectively, and unsuccessful action attempts also consume the budget.

\clearpage

\subsection{RoboCasa}
\label{app:robocasa}

\subsubsection{Benchmark Construction and Task Setup}
\label{app:robocasa_benchmark}

RoboCasa provides long-horizon household manipulation tasks together with structured task annotations, successful trajectories, and simulator observations.
While the original benchmark focuses on robotic manipulation at the execution level, we adapt its Target Composite tasks to evaluate visual high-level planning.
Specifically, we convert the annotated manipulation trajectories into high-level planning sequences by abstracting low-level robot control into task-level skill stages and using the corresponding simulator observations as intermediate visual states.

\subsubsection{Planning Interface and Execution Protocol}
\label{app:robocasa_interface}

We abstract RoboCasa's low-level robot control into a unified high-level action interface consisting of eight task-level APIs: \texttt{NAVIGATE}, \texttt{PICK}, \texttt{PLACE}, \texttt{OPEN}, \texttt{CLOSE}, \texttt{PRESS}, \texttt{ROTATE}, and \texttt{TILT}.
These actions abstract away low-level motion and manipulation details, allowing the planner to operate directly over task-level manipulation steps.
All compared methods receive the same task description, RGB observation, public object and fixture vocabulary, high-level action API, and executed-action history, while hidden simulator states and reference action sequences are not exposed.

We evaluate two complementary planning protocols: \textbf{one-shot planning} and \textbf{closed-loop replanning}.
In the one-shot setting, the planner generates a complete sequence of high-level actions from the initial observation and task description before execution.
In the replanning setting, the planner predicts one high-level action at a time. After execution, the resulting RGB observation and SUCCESS/FAILURE feedback are returned to the planner for the next decision.
This protocol allows the planner to adapt its subsequent actions to the actual state reached during execution.

\subsubsection{Evaluation Protocol}
\label{app:robocasa_evaluation}

We evaluate predicted high-level plans by replaying them against the official RoboCasa action boundaries while preserving the corresponding simulator states and task goal checker.
At each boundary, a predicted action is considered \textbf{SUCCESS} if it matches a valid recorded transition from the current state and advances execution to the corresponding next state.
A well-formed action that is not applicable at the current boundary is marked as \textbf{FAILURE}.
If a predicted action cannot be resolved because the available official trajectory does not contain the corresponding branch, we mark it as UNKNOWN rather than treating it as an ordinary planning failure.
A task is counted as successful only when execution reaches the official task goal condition.

We begin with a fixed set of 100 held-out test cases.
For each planning protocol, we evaluate all methods on the shared subset of cases for which no method encounters an UNKNOWN transition.
For one-shot planning, the generated plan is limited to at most 40 high-level actions.
For closed-loop replanning, the planner is allowed at most 25 environment interaction attempts, with each subsequent decision conditioned on the latest RGB observation and execution history.

\subsubsection{Representative High-Level Planning Trajectories}
\label{app:robocasa_trajectory}

\begin{figure}[H]
    \centering

    \includegraphics[width=0.98\textwidth]{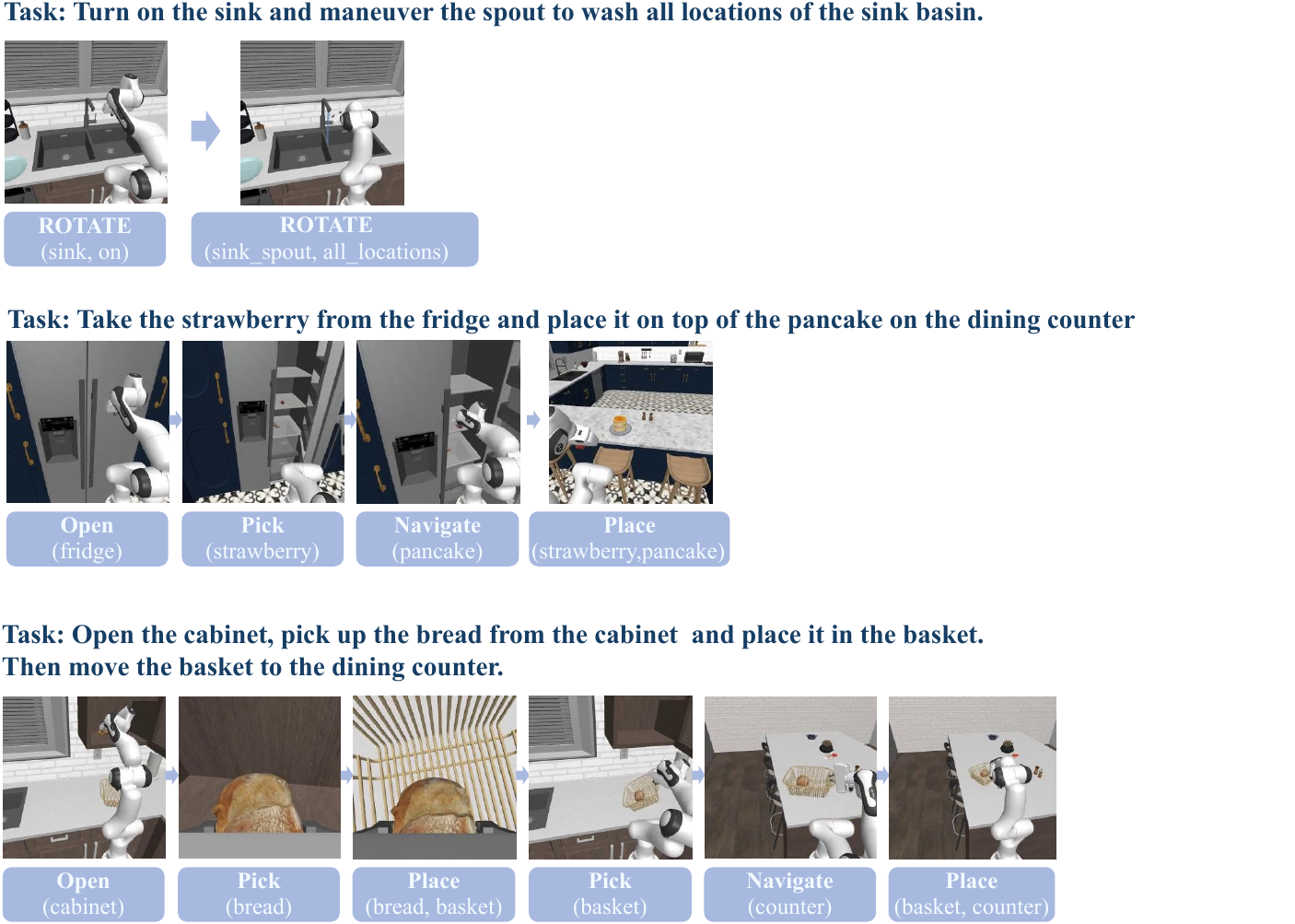}

    \vspace{6pt}

    \includegraphics[width=0.98\textwidth]{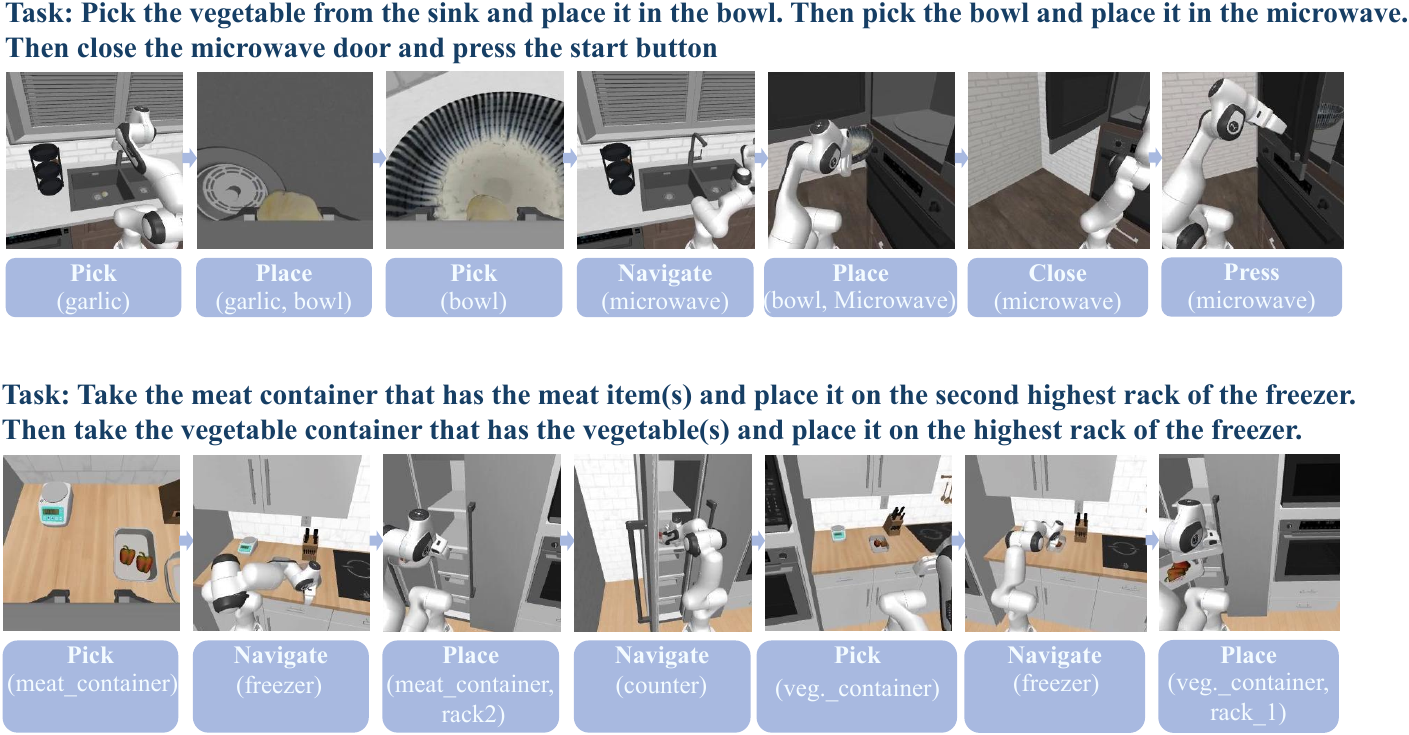}

    \caption{\textbf{Representative high-level planning trajectories in RoboCasa.}
    We show five representative tasks covering object manipulation, navigation, container interaction, fixture operation, and multi-step composition.
    Each trajectory is represented by intermediate RGB observations and the corresponding high-level actions from our unified action interface.}
    \label{fig:robocasa_trajectory}
\end{figure}

Figure~\ref{fig:robocasa_trajectory} illustrates how an official RoboCasa manipulation trajectory is represented under our high-level planning interface.
Each intermediate RGB observation corresponds to a task-level action boundary, while the associated low-level robot control is abstracted into the unified high-level action API.
This representation preserves the original task progression while exposing only the action granularity required for high-level planning.

\clearpage

\subsection{SWITCH}
\label{app:switch}

We use SWITCH without modifying its original task formulation or evaluation protocol.
We select four subtasks that most directly evaluate action-conditioned state-transition reasoning:~\textbf{Action},~\textbf{Recovery},~\textbf{Verification Action}, and~\textbf{Verification State}.
These subtasks respectively probe action selection, recovery from interaction failures, verification of action outcomes, and reasoning about the resulting environment state.
We follow the original input and label formats and report prediction accuracy for each subtask as well as the overall average.

\begin{table}[H]
\centering
\caption{\textbf{Statistics of the selected SWITCH subtasks.}
We evaluate four interaction-reasoning categories from the official SWITCH benchmark.}
\label{tab:switch_statistics}
\setlength{\tabcolsep}{8pt}
\begin{tabular}{lc}
\toprule
Subtask & \# Examples \\
\midrule
Action               & 442 \\
Recovery             & 38  \\
Verification Action  & 63  \\
Verification State   & 225 \\
\midrule
Total                & 768 \\
\bottomrule
\end{tabular}
\end{table}

We report prediction accuracy for each subtask and use the macro-average over the four subtasks as the overall score.
All methods are evaluated on the same examples using the original SWITCH input and label formats.

\begin{figure}[H]
    \centering
    \includegraphics[width=\textwidth]{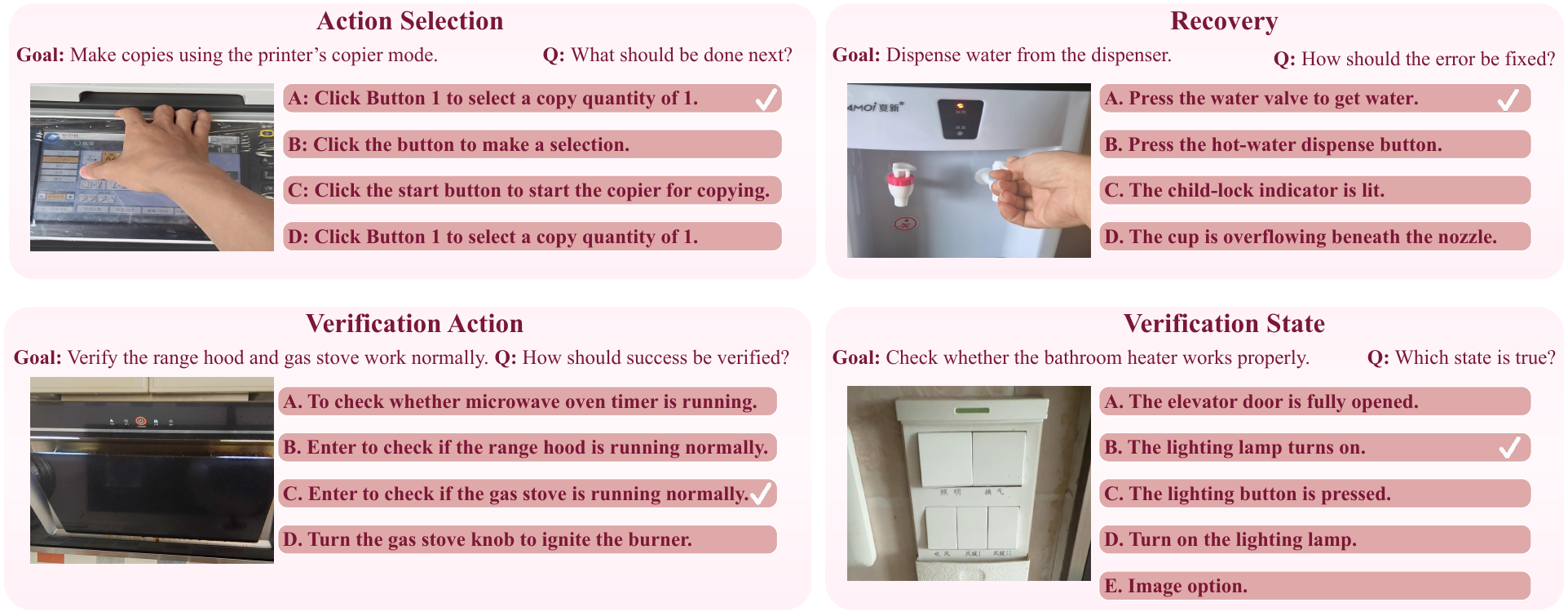}
    \caption{\textbf{Representative cases from the selected SWITCH subtasks.}
    We show examples of Action Selection, Recovery, Verification Action, and Verification State.
    These subtasks respectively evaluate next-action prediction, error recovery, active verification, and state verification under embodied interactions.}
    \label{fig:switch_cases}
\end{figure}

\textbf{Representative Cases.}
Figure~\ref{fig:switch_cases} presents representative examples from the four selected SWITCH subtasks, illustrating the distinct interaction-reasoning capabilities evaluated by each category.

\clearpage

\section{Additional Details of CausGround}
\label{app:method}
\subsection{Factorized World-State Representation}
\label{app:factorized_state}

\begin{figure}[H]
    \centering
    \includegraphics[width=\textwidth]{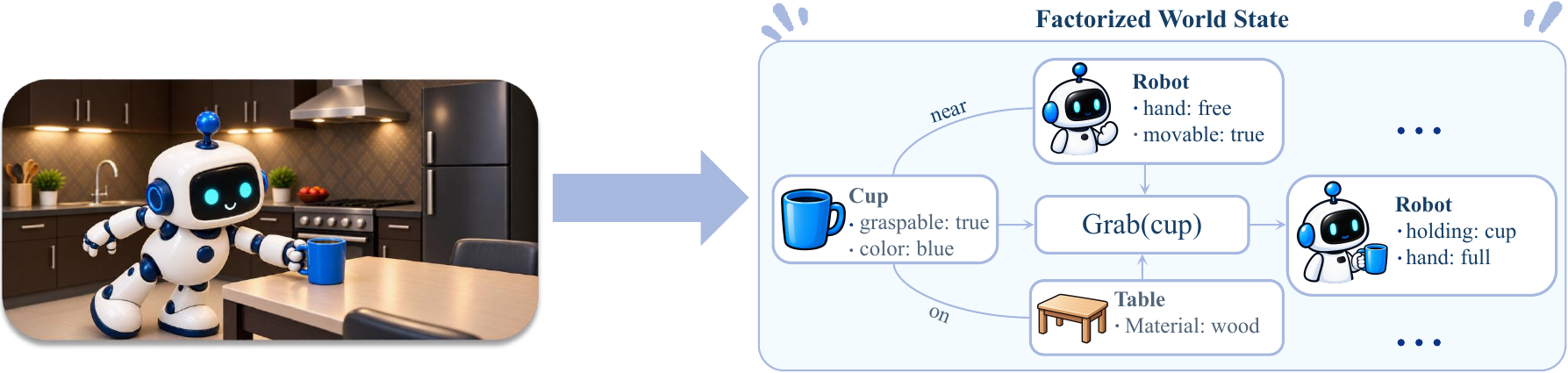}
    \caption{\textbf{Illustration of the factorized world state.}
    Visual observations are converted into a structured representation that explicitly separates object attributes from inter-object relations.
    The same representation is used to describe the current state as well as the preconditions and effects associated with environment interactions.}
    \label{fig:factorized_world_state}
\end{figure}

Figure~\ref{fig:factorized_world_state} provides an intuitive example of the factorized world-state representation used throughout CausGround.
Given a visual observation, each perceived object is represented by its semantic category together with explicit attribute--value pairs, while interactions among objects are represented as typed relations.
For example, the cup can be described by attributes such as \texttt{graspable=true} and \texttt{color=blue}, while its relation to the robot and table is represented through \texttt{NEAR(robot,cup)} and \texttt{ON(cup,table)}.
The resulting structured state provides a shared representation for describing the current environment, action preconditions, and action effects.

Multiple objects of the same semantic category remain separate instances in the world-state graph and can be distinguished when necessary through their observed attributes and relational context.
Relations preserve their original semantics. For example, \texttt{CLOSE(x,y)} is not interpreted as \texttt{ON(x,y)} or \texttt{INSIDE(x,y)} without supporting evidence.
This representation allows CausGround to identify individual state factors involved in action execution and state transitions, which are further used for causal knowledge construction in Sec.~\ref{app:causal_construction}.

\subsection{Action-Centered Causal Knowledge Construction}
\label{app:causal_construction}

\begin{figure}[H]
    \centering
    \includegraphics[width=\textwidth]{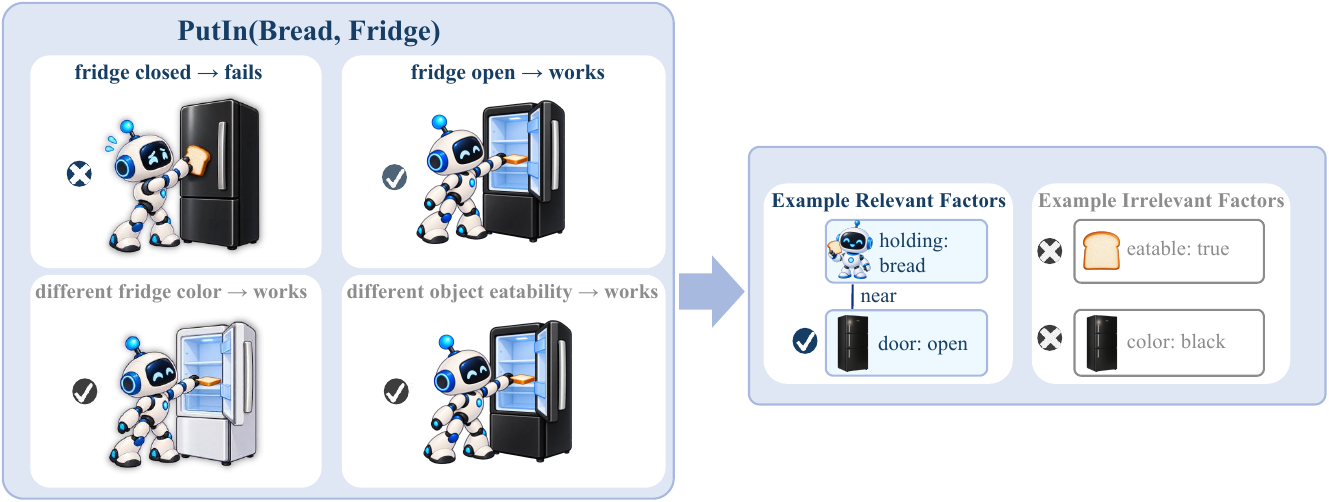}
    \caption{\textbf{Illustration of action-centered causal factor identification.}
    CausGround compares executions of the same grounded action under variations of pre-action state factors.
    Factors associated with changes in action executability are retained as relevant preconditions, whereas incidental factors that do not affect the execution outcome are discarded.}
    \label{fig:causal_factor_identification}
\end{figure}

CausGround constructs reusable causal knowledge at the level of individual actions.
For each grounded action, we collect interaction instances with different pre-action world states and compare their execution outcomes.
As illustrated in Fig.~\ref{fig:causal_factor_identification}, factors whose variation is consistently associated with whether the action can be successfully executed are retained as action-relevant preconditions, while factors that vary without affecting executability are treated as incidental and discarded.
For example, \texttt{PUTIN(bread,fridge)} fails when the fridge is closed but succeeds when it is open, indicating that the fridge state is relevant to action execution.
In contrast, changing unrelated properties such as the fridge color does not affect the outcome and therefore should not be retained as a precondition.

Action effects are extracted by comparing the pre and post-action world states of successful executions.
State changes that consistently follow an action are retained as its effects, while unrelated or non-repeatable changes are removed.
Together, the extracted preconditions and effects describe both what must hold before an action can be executed and how the world is expected to change after execution.

We further organize this knowledge at two levels of specificity.
Across grounded instances of the same action family, preconditions and effects that remain valid across different objects are abstracted as \textbf{generic knowledge}, while regularities that are specific to particular objects or fixtures are retained as \textbf{instance-specific knowledge}.
For example, \texttt{OPEN(fridge)} and \texttt{OPEN(microwave)} may share the generic effect that the target becomes open, while preserving different object-specific requirements when necessary.

Finally, CausGround connects individual actions through effect--precondition dependencies.
When an effect produced by one action satisfies a precondition required by another action, a directed dependency is introduced between them.
Such dependencies allow reusable single-action knowledge to be composed into multi-step causal structures, which are subsequently retrieved and grounded according to the current task in Sec.~\ref{app:task_grounding}.

\subsection{Task-Grounded Planning}
\label{app:task_grounding}

\begin{figure}[H]
    \centering
    \includegraphics[width=0.75\textwidth]{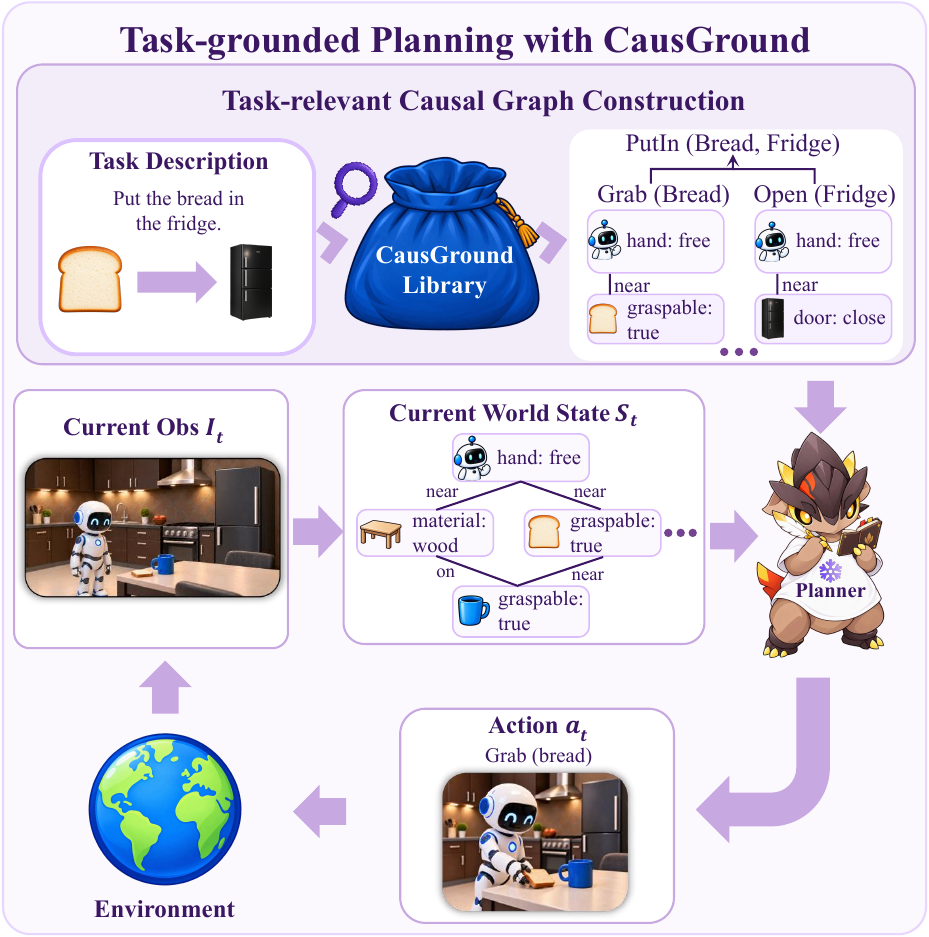}
    \caption{\textbf{Task-grounded planning in CausGround.}
    The task description is used to retrieve task-relevant causal knowledge from the CausGround library.
    At each planning step, the current observation is converted into a factorized world state and used to ground the retrieved causal graph in the current environment state before action prediction.
    The predicted action is executed in the environment, and the resulting observation is returned for the next decision step.}
    \label{fig:task_grounded_planning_appendix}
\end{figure}

Figure~\ref{fig:task_grounded_planning_appendix} illustrates the task-grounded planning procedure in CausGround.
Given a task description, CausGround first retrieves task-relevant causal knowledge from the interaction library and organizes it into a task-level causal graph.
At each decision step, the current visual observation is converted into a factorized world state, which is then used to ground the retrieved causal knowledge in the current environment state before being passed to the planner.
The planner predicts the next high-level action, the environment executes it, and the resulting observation is fed back for the next planning step.

More concretely, the retrieved task knowledge contains actions, preconditions, effects, and effect--precondition dependencies that are potentially useful for achieving the task goal.
The current world state is used to identify which retrieved factors are already satisfied, which are currently violated, and which candidate actions are executable under the present state.
This grounded task knowledge is provided together with the current observation to support state-aware high-level planning.
\clearpage

\section{Implementation Details and Prompts}
\label{app:implementation}

\subsection{Common Experimental Setup}
\label{app:common_setup}

For all comparisons, we use matched experimental settings across methods.
Specifically, methods evaluated within the same benchmark share the same planner backbone, task descriptions, visual observations, high-level action space, and evaluation protocol.
All planner parameters remain frozen throughout evaluation, such that performance differences arise from the planning context and method-specific reasoning mechanisms rather than parameter updates.

In the main experiments, we instantiate the planner with three Qwen3.5 backbones: Qwen3.5-9B, Qwen3.5-27B, and Qwen3.5-35B-A3B.
For the cross-family analysis in Fig.~\ref{fig:model_generalization}, we additionally evaluate Gemma 4 E4B, Gemma 4 12B, InternVL3.5-8B, and InternVL3.5-38B under the same VirtualHome planning interface and evaluation protocol.
For VirtualHome and RoboCasa, all methods operate on the same visually grounded high-level planning interface defined in Appendix~\ref{app:environment}.
For SWITCH, all methods are evaluated on the same selected examples using the original benchmark input and label formats.

\subsection{Baseline Implementations}
\label{app:baselines}

Within each environment, all methods share the same task descriptions, visual observations, public action, object interface, and evaluation protocol.

\textbf{Vanilla Method.}
At each step, the planner directly predicts one high-level semantic action from the current RGB observation, task instruction, public action/object vocabulary, and actual execution history.
It does not use the CausGround, a dynamic local world state, self-revision, or an auxiliary graph-planning module.
The exact planner interface is shown in Appendix~\ref{app:prompts}.

\textbf{SelfReVision.}
We adapt the official SelfReVision plan--critique--revision pipeline to the shared VirtualHome and RoboCasa semantic-action interface.
For each decision, the model generates an initial multi-step plan, critiques it, and produces a revised plan; subsequent refinements are selected using the plan-comparison verifier.
We use a maximum refinement depth of three rounds.
Only the first action of the selected plan is executed in the environment, after which planning is repeated from the updated observation.
The original SelfReVision prompt family and our semantic-action adaptation are provided in Appendix~\ref{app:prompts}.

\textbf{LookPlanGraph.}
We retain the graph-planning and SceneSimulator(SGS) components of LookPlanGraph and introduce a VirtualHome and RoboCasa compatibility layer.
At each step, the current RGB observation is converted into a public visual graph, which is conservatively merged with the method's memory before graph-based planning and SGS verification.
Only the first proposed action is executed, and graph-state updates are committed only after real executor success.
The graph is constructed from visual observations.
Its perception, graph-formatting, and planning prompts are provided in Appendix~\ref{app:prompts}.

\subsection{CausGround Implementation}
\label{app:causground_impl}

\subsubsection{CausGround Library Construction}
\label{app:library_construction}
We construct the CausGround library from environment interaction experience, using different causal identification protocols according to the intervention capability of each environment.

For VirtualHome, we conduct controlled interaction experiments for each action by actively varying candidate pre-action state factors. A factor is retained as a precondition only when violating it prevents successful execution, while action effects are identified from state changes that consistently follow successful executions.
This controlled setting allows us to directly test which state factors are necessary for an action and which changes reliably result from its execution.
The resulting global VirtualHome library contains 58,458 semantic action-hyperedge records, which are consolidated into 5,614 normalized action rules covering 40 action families.
These rules span 193 semantic node types, 28 state-attribute types, and six explicit relation types, and contain 113,847 required-condition records and 154,597 effect records.

For environments where controlled interventions are less practical (e.g., RoboCasa and SWITCH), we instead identify causal regularities under a strict consistency criterion from repeated official interaction trajectories.
A candidate precondition is retained only when it is consistently supported across successful executions; if the action succeeds even once without that factor, the candidate is excluded.
Candidate effects are filtered similarly, retaining only state changes that consistently follow successful executions.
The resulting RoboCasa CausGround library contains 98 grounded action mechanisms covering 8 high-level action families and 225 semantic object and fixture types, with 254 condition records and 416 effect records.
For SWITCH, we construct the causal knowledge from its interaction trajectories using the same consistency-based extraction procedure and apply the resulting knowledge to the selected interaction-reasoning subtasks.

\subsubsection{Task-Time Retrieval and Grounding}
\label{app:retrieval_grounding}

Given a task description, CausGround identifies the corresponding goal conditions and required interaction events, and retrieves a task-relevant causal subgraph from the environment-specific library.
The retrieved knowledge contains relevant action conditions, effects, and producer dependencies, and is loaded once at the beginning of each episode rather than reconstructed at every planning step.

During execution, the retrieved causal knowledge is repeatedly \textbf{grounded} in the current environment state.
At each step, the current RGB observation is converted into the factorized local world state, which is used to determine which retrieved conditions are already satisfied and which prerequisite actions remain relevant.
The planner first predicts a semantic high-level action, such as
\texttt{OPEN(fridge)}. Its semantic arguments are then matched to the
corresponding object instances represented in the current world state before the action is sent to the executor.
After execution, the new observation and public success/failure history update the local world state for the next decision, while the retrieved causal subgraph itself remains fixed.

\subsection{Prompt Templates}
\label{app:prompts}
For reproducibility, we provide the core prompt templates used by all evaluated methods. Environment-specific action and object vocabularies are dynamically inserted into the corresponding templates at runtime.

\subsubsection{Vanilla}

\begin{vanillabox}{Vanilla Planner Prompt}

Choose exactly one next household action from the current image and instruction.

Return only the following JSON format:

\begin{verbatim}
{
  "action": "ACTION",
  "arguments": [
    "semantic object class"
  ]
}
\end{verbatim}

Do not output simulator IDs, a plan, or an action sequence.

\textbf{Task:} \texttt{<task instruction>}

\textbf{Action Signatures:} \texttt{<public action signatures>}

\textbf{Object Vocabulary:} \texttt{<public object vocabulary>}

\textbf{History:} \texttt{<public execution history>}

\end{vanillabox}

\subsubsection{SelfReVision}

\begin{selfrevisionbox}{Initial Plan Prompt}

You are writing instructions for a robot based on the current visual observation.
Make a detailed plan that responds to the user's request and is specific to the current environment.

Use only information grounded in the current observation. Unseen objects or relations should remain unknown.
Do not infer hidden states or simulator-specific IDs.
Use only the provided environment-specific action and object vocabularies.

Return only the following JSON format:

\begin{verbatim}
{
  "Plan": [
    {
      "action": "ACTION",
      "arguments": [
        "object"
      ],
      "brief_reason": "short reason"
    }
  ]
}
\end{verbatim}

Only the first action of the selected plan will be executed in the environment.

\textbf{Task:} \texttt{<task instruction>}

\textbf{Action Signatures:} \texttt{<public action signatures>}

\textbf{Object Vocabulary:} \texttt{<public object vocabulary>}

\end{selfrevisionbox}

\clearpage
\begin{selfrevisionbox}{Criticism Prompt}

You are reviewing a high-level plan for a robot based on the user's request and the current visual observation.

Identify critical flaws, gaps, or missed opportunities that materially affect the feasibility, clarity, or alignment of the plan.
Ground your feedback in the visual context and user intent.
Avoid minor stylistic issues.

Return only a clean numbered list enclosed in:

\begin{verbatim}
<critic>
1. ...
2. ...
</critic>
\end{verbatim}

\textbf{Task:} \texttt{<task instruction>}

\textbf{Current Plan:} \texttt{<current JSON plan>}

\end{selfrevisionbox}

\begin{selfrevisionbox}{Revision Prompt}

You are revising a high-level robot plan based on critical feedback, the user's request, and the current visual observation.

Make meaningful changes that address the identified issues. Prioritize feasibility and visual grounding, omit redundant details, and avoid unsupported speculation.

Return only the revised plan in the following JSON format:

\begin{verbatim}
{
  "Plan": [
    {
      "action": "ACTION",
      "arguments": [
        "object"
      ],
      "brief_reason": "short reason"
    }
  ]
}
\end{verbatim}

Use only the provided environment-specific action and object vocabularies.
Only the first action of the selected revised plan will be executed.

\textbf{Task:} \texttt{<task instruction>}

\textbf{Current Plan:} \texttt{<current JSON plan>}

\textbf{Feedback:} \texttt{<critic feedback>}

\end{selfrevisionbox}

\begin{selfrevisionbox}{Plan Verifier Prompt}

You are evaluating two high-level plans for a robot based on the user's request and the current visual observation.

Respond with \texttt{"yes"} if Plan 2 better fulfills the user's request; otherwise respond with \texttt{"no"}.

Judge the two plans based on visual grounding, logical order, completeness, coverage, and task alignment.

Return only:

\begin{verbatim}
yes
\end{verbatim}

or

\begin{verbatim}
no
\end{verbatim}

\textbf{Task:} \texttt{<task instruction>}

\textbf{Plan 1:} \texttt{<current JSON plan>}

\textbf{Plan 2:} \texttt{<revised JSON plan>}

\end{selfrevisionbox}

\subsubsection{LookPlanGraph}

\begin{lookplangraphbox}{Scene Description Prompt}

Describe the current visible scene using the provided object and environment vocabulary.

Report only information that is visually grounded in the current observation.
Include visible objects, furniture, appliances, and their clearly observable states or relations.

Do not infer hidden contents, unobserved relations, simulator-specific IDs, or task completion.
Objects that are occluded or not visible should remain unknown.

Use only semantic names from the provided vocabulary.

\textbf{Task:} \texttt{<task instruction>}

\textbf{Object Vocabulary:} \texttt{<public object vocabulary>}

\end{lookplangraphbox}

\begin{lookplangraphbox}{Graph Formatting Prompt}

Convert the scene description into a structured public visual graph.

Return a JSON list in the following format:

\begin{verbatim}
[
  {
    "name": "class-1",
    "type": "asset|object",
    "states": [
      "state"
    ],
    "relation": "ontop_of|inside_of|null",
    "related_to": "class-1|null",
    "affordances": [
      "affordance"
    ]
  }
]
\end{verbatim}

Include all objects that are explicitly described in the current scene.
Use observable states such as \texttt{open}, \texttt{closed}, \texttt{on}, or \texttt{off} only when supported by the visual description.

Do not invent objects, states, or relations.
Reuse previously observed semantic handles when the object identity is consistent.
Objects that are not currently observed should not be treated as nonexistent.

\textbf{Scene Description:} \texttt{<current scene description>}

\textbf{Object Vocabulary:} \texttt{<public object vocabulary>}

\end{lookplangraphbox}

\begin{lookplangraphbox}{Planning and SGS Prompt}

Given the current public visual graph, task instruction, and execution history,
select the next high-level action that best advances the task.

Use only information contained in the public graph and current observation.
Unobserved objects, states, or relations should remain unknown.
Do not infer hidden contents or simulator-specific IDs.

Actions are not assumed to succeed automatically.
Only confirmed execution success may update the stored graph state.
Failed actions should remain in the execution history without committing their predicted effects.

Return only one semantic high-level action in the following JSON format:

\begin{verbatim}
{
  "action": "ACTION",
  "arguments": [
    "object"
  ]
}
\end{verbatim}

\textbf{Task:} \texttt{<task instruction>}

\textbf{Public Visual Graph:} \texttt{<relevant graph nodes and relations>}

\textbf{Action Signatures:} \texttt{<public action signatures>}

\textbf{Object Vocabulary:} \texttt{<public object vocabulary>}

\textbf{Execution History:} \texttt{<public success/failure history>}

\textbf{Held Objects / Agent State:} \texttt{<public agent state>}

\end{lookplangraphbox}

\subsubsection{CausGround}

\begin{causgroundbox}{CausGround Planner Prompt}

Given the current visual observation, task instruction, factorized world state, and retrieved task-relevant causal knowledge, select the next high-level action that best advances the task.

Use the current world state to determine which action preconditions are already satisfied and which prerequisites remain unmet.

Use the retrieved action effects and producer dependencies to identify actions that can establish missing prerequisites before attempting downstream actions.

Treat the causal knowledge as a task-relevant reference rather than a fixed action sequence.

Unknown or unobserved state factors should remain unknown.

Do not infer simulator-specific IDs, private executor states, or reference trajectories.

Return only one semantic high-level action in the following JSON format:

\begin{verbatim}
{
  "action": "ACTION",
  "arguments": [
    "object"
  ]
}
\end{verbatim}

\textbf{Task:} \texttt{<task instruction>}

\textbf{Current World State:}

\texttt{<factorized local world state>}

\textbf{Retrieved Causal Knowledge:}

\texttt{<action conditions, effects, and producer dependencies>}

\textbf{Required Task Events:}

\texttt{<task-relevant required events>}

\textbf{Action Signatures:}

\texttt{<public action signatures>}

\textbf{Object Vocabulary:}

\texttt{<public object vocabulary>}

\textbf{Execution History:}

\texttt{<public success/failure history>}

\end{causgroundbox}

\section{Additional Ablation Studies}
\label{app:additional_ablations}

\subsection{Causal Knowledge vs. Factorized State Across Backbones}
\label{app:ablation_backbones}

The main paper reports the component comparison using Qwen3.5-27B. Here, we extend the same controlled comparison to Qwen3.5-9B and Qwen3.5-35B-A3B, while retaining the 27B result for reference.

\begin{table}[H]
\centering
\caption{\textbf{Component ablations across planner backbones.} Contributions of the factorized state and causal interaction knowledge across different Qwen3.5 backbones.}
\label{tab:ablation_backbones}
\setlength{\tabcolsep}{3.5pt}
\renewcommand{\arraystretch}{1.08}

\begin{tabular}{lcc|cc|cc}
\toprule
\multirow{2}{*}{Method}
& \multicolumn{2}{c|}{Qwen3.5-9B}
& \multicolumn{2}{c|}{Qwen3.5-27B}
& \multicolumn{2}{c}{Qwen3.5-35B-A3B} \\
\cmidrule(lr){2-3}
\cmidrule(lr){4-5}
\cmidrule(lr){6-7}
& Success Rate & Progress
& Success Rate & Progress
& Success Rate & Progress \\
\midrule

Vanilla
& 10.7 & 23.8
& 33.3 & 46.9
& 30.7 & 44.4 \\

+ Factorized State
& 17.3 & 32.3
& 36.0 & 55.1
& 40.0 & 56.3 \\

+ Causal Knowledge
& 41.3 & 57.6
& 53.3 & 70.8
& 50.7 & 69.1 \\

\rowcolor{oursblue}
CausGround (Ours)
& \textbf{44.0} & \textbf{61.5}
& \textbf{62.7} & \textbf{79.8}
& \textbf{53.3} & \textbf{72.6} \\

\bottomrule
\end{tabular}
\end{table}

The same trend is observed across all three planner backbones. Both components consistently improve over Vanilla, with causal interaction knowledge providing the larger individual gain. 
Their combination achieves the highest success rate and final progress for every backbone, confirming that the factorized state and causal knowledge are complementary.
This extends the 27B component analysis in the main paper to different planner scales.

\subsection{Causal vs. Semantic Retrieval}
\label{app:ablation_retrieval}

We further examine the contribution of the two retrieval paths in CausGround using Qwen3.5-9B.
We compare semantic retrieval alone, causal backward retrieval alone, and their combination, while keeping the CausGround pipeline, planner backbone, observations, grounding, and execution protocol unchanged.

\begin{table}[H]
\centering
\caption{\textbf{Ablation of retrieval strategies.}
We compare semantic retrieval, causal backward retrieval, and their combination using Qwen3.5-9B on the VirtualHome benchmark.}
\label{tab:ablation_retrieval}
\setlength{\tabcolsep}{10pt}
\renewcommand{\arraystretch}{1.08}

\begin{tabular}{lcc}
\toprule
Retrieval Strategy & Success Rate & Progress \\
\midrule
Vanilla
& 10.7 & 23.8 \\

+ Semantic Retrieval
& 18.7 & 37.6 \\

+ Causal Retrieval
& 42.7 & 60.4 \\

\rowcolor{oursblue}
CausGround (Ours)
& \textbf{44.0} & \textbf{61.5} \\
\bottomrule
\end{tabular}
\end{table}

The results show that causal backward retrieval is the primary contributor to CausGround's retrieval mechanism, consistent with our method design. By exposing action preconditions and effects, CausGround helps the planner reason about when an action is executable and how it changes the world, thereby improving embodied planning.
Semantic retrieval alone improves success rate from 10.7\% to 18.7\% and progress from 23.8\% to 37.6\%, while causal retrieval yields substantially larger gains, reaching 42.7\%  success rate and 60.4\% progress.
Combining the two retrieval paths further improves performance to 44.0\%  success rate and 61.5\% progress, indicating that causal retrieval accounts for most of the planning gain.

\subsection{Action Preconditions vs. Effects}
\label{app:ablation_condition_effect}

We further isolate the contributions of action preconditions and effects using Qwen3.5-9B.
Starting from the same CausGround pipeline, we construct two variants that expose only action preconditions or only action effects, while keeping the planner backbone, retrieval procedure, observations, grounding, and execution protocol unchanged.

\begin{table}[H]
\centering
\caption{\textbf{Ablation of action preconditions and effects.}
We separately expose only action preconditions or only action effects to evaluate their individual contributions on the VirtualHome benchmark.}
\label{tab:ablation_condition_effect}
\setlength{\tabcolsep}{10pt}
\renewcommand{\arraystretch}{1.08}

\begin{tabular}{lcc}
\toprule
Knowledge Variant & Success Rate & Progress \\
\midrule
Vanilla
& 10.7 & 23.8 \\

+ Preconditions Only
& 36.0 & 52.1 \\

+ Effects Only
& 38.7 & 55.3 \\

\rowcolor{oursblue}
CausGround (Ours)
& \textbf{44.0} & \textbf{61.5} \\
\bottomrule
\end{tabular}
\end{table}

Both action preconditions and effects independently provide substantial gains over Vanilla. Preconditions help the planner determine when an action is executable, while effects describe how the world changes after execution.
Effects alone achieve slightly higher performance than preconditions alone, but combining both yields the best result, reaching 44.0\% success rate and 61.5\% progress.
These results support the design of CausGround, where preconditions and effects provide complementary causal information for embodied planning.

\clearpage

\section{Real-World Deployment}
\label{app:real_world}

We further evaluate CausGround on a dual-arm robot in a real-world tabletop household manipulation setting.
In this deployment, the MLLM is used only to interpret the natural-language instruction into a goal state.
Starting from the interpreted goal, CausGround performs backward reasoning over action effects and preconditions to identify the prerequisite actions required for execution and construct a high-level action sequence.
The resulting semantic actions are then carried out by the robot's existing low-level controller.

\begin{figure}[H]
    \centering
    \includegraphics[width=\textwidth]{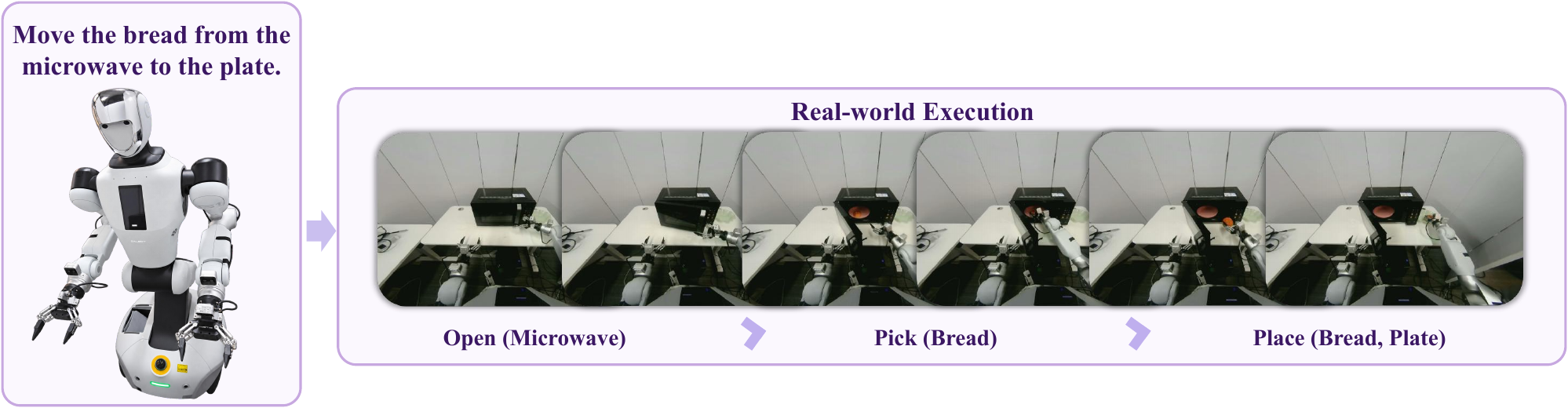}
    \caption{\textbf{Real-world execution of the microwave task.}
    Given the instruction ``Move the bread from the microwave to the plate,'' CausGround produces the high-level sequence~\texttt{OPEN(Microwave)} $\rightarrow$~\texttt{PICK(Bread)} $\rightarrow$~\texttt{PLACE(Bread, Plate)}, which is executed by the robot in the real environment.}
    \label{fig:real_microwave}
\end{figure}

\begin{figure}[H]
    \centering
    \includegraphics[width=\textwidth]{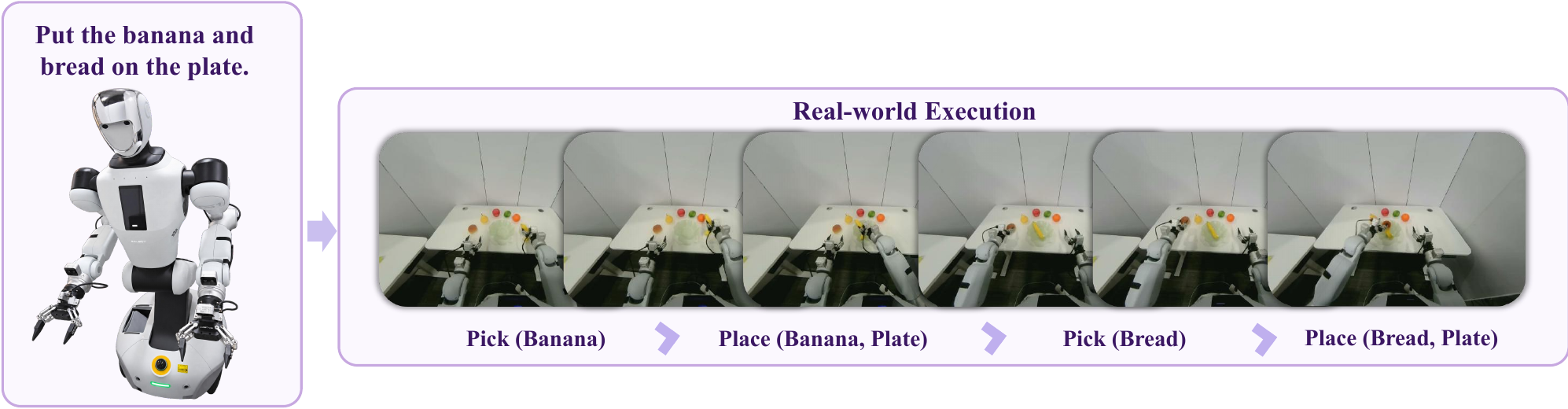}
    \caption{\textbf{Real-world execution of the multi-object task.} Given the instruction ``Put the banana and bread on the plate,'' CausGround sequentially generates~\texttt{PICK(Banana)} $\rightarrow$~\texttt{PLACE(Banana, Plate)} $\rightarrow$~\texttt{PICK(Bread)} $\rightarrow$~\texttt{PLACE(Bread, Plate)}, demonstrating multi-object sequential manipulation.}
    \label{fig:real_multiobject}
\end{figure}

The two tasks illustrate how CausGround derives executable high-level actions from causal dependencies without relying on an additional MLLM planner for high-level action selection.
In the microwave task, backward reasoning identifies the intermediate retrieval requirement and the prerequisite needed to access the bread before execution.
In the multi-object task, backward reasoning decomposes the joint goal into object-wise manipulation requirements and identifies the prerequisite interactions needed to place both objects on the plate.